\documentclass[10pt,journal,compsoc]{IEEEtran}
\usepackage{graphicx}
\usepackage{subfigure}  
\usepackage{times}
\usepackage{epsfig}
\usepackage{graphicx}
\usepackage{amssymb}
\usepackage{amsmath}
\usepackage{amsthm}
\usepackage{bm}
\usepackage{subfigure}
\usepackage{booktabs}
\usepackage{multirow}
\usepackage{float}
\usepackage[utf8]{inputenc}
\newcommand{\comment}[1]{}

\usepackage{overpic}
\usepackage{caption2}

\newtheorem{theorem}{Theorem}

\usepackage{overpic}

\ifCLASSOPTIONcompsoc
  \usepackage[nocompress]{cite}
\else
  \usepackage{cite}
\fi
\usepackage{algorithmic}
\begin{document}
%
\title{Cluster-Level Sparse Multi-Instance Learning for Whole-Slide Images}
%
%
%
%

\author{
Yuedi~Zhang,~\IEEEmembership{Student Member,~IEEE}
Zhixiang~Xia,~\IEEEmembership{Student Member,~IEEE}
Guosheng~Yin,~\IEEEmembership{Senior Member,~IEEE}and~Bin~Liu,~\IEEEmembership{Member,~IEEE}
\IEEEcompsocitemizethanks{\IEEEcompsocthanksitem M. Shell was with the Department
of Electrical and Computer Engineering, Georgia Institute of Technology, Atlanta,
GA, 30332.\protect\\
E-mail: see http://www.michaelshell.org/contact.html
\IEEEcompsocthanksitem J. Doe and J. Doe are with Anonymous University.}
\thanks{Manuscript received April 19, 2005; revised August 26, 2015.}}

%
%

\markboth{Journal of \LaTeX\ Class Files,~Vol.~14, No.~8, August~2015}%
{Shell \MakeLowercase{\textit{et al.}}: Bare Demo of IEEEtran.cls for Computer Society Journals}
%



\IEEEtitleabstractindextext{%
\begin{abstract}
Multi-Instance Learning (MIL) is pivotal for analyzing complex, weakly labeled datasets, such as whole-slide images (WSIs) in computational pathology, where bags comprise unordered collections of instances with sparse diagnostic relevance. Traditional MIL approaches, including early statistical methods and recent attention-based frameworks, struggle with instance redundancy and lack explicit mechanisms for discarding non-informative instances, limiting their robustness and interpretability. We propose Cluster-level Sparse MIL (csMIL), a novel framework that integrates global-local instance clustering, within-cluster attention, and cluster-level sparsity induction to address these challenges. Our csMIL first performs global clustering across all bags to establish $K$ cluster centers, followed by local clustering within each bag to assign cluster labels. Attention scores are computed within each cluster, and sparse regularization is applied to cluster weights, enabling the selective retention of diagnostically relevant clusters while discarding irrelevant ones. This approach enhances robustness to noisy instances, improves interpretability by identifying critical regions, and reduces computational complexity. Theoretical analysis demonstrates that csMIL requires $O(s \log K)$ bags to recover $s$ relevant clusters, aligning with compressed sensing principles. Empirically, csMIL achieves state-of-the-art performance on two public histopathology benchmarks (CAMELYON16, TCGA-NSCLC).
 \end{abstract}

}

\maketitle

\IEEEdisplaynontitleabstractindextext

%
\IEEEpeerreviewmaketitle

\IEEEraisesectionheading{\section{Introduction}\label{sec:introduction}}

%
%
%
%
\IEEEPARstart {T}{he} paradigm of multi-instance learning (MIL) has become indispensable in analyzing complex, weakly labeled datasets where data objects (e.g., histopathology slides, satellite images) are represented as unordered collections of instances. The standard MIL assumption posits that a bag is positive if at least one instance is positive, and negative if all instances are negative.
In medical imaging, for example, a whole-slide image (bag) may contain thousands of tissue patches (instances), yet only a sparse subset correlates with diagnostic labels such as tumor malignancy. MIL has become a cornerstone in medical imaging, especially for whole-slide image (WSI) analysis \cite{lu2021data,GADERMAYR2024102337,liu2024pseudo,liu2024advmil}.

Early MIL methods relied on geometric and statistical approaches. For instance, Maron and Lozano-Pérez (1998) \cite{maron1997framework} proposed Diverse Density, a probabilistic framework to identify regions in feature space with high positive instance density. Andrews et al. (2002) introduced mi-SVM and MI-SVM \cite{andrews2002support}, adapting support vector machines to MIL by treating instance labels as latent variables or optimizing bag-level margins, respectively.
Toward the decade’s end, neural networks began influencing MIL. Sun et al. (2016) explored convolutional neural networks for MIL in image classification, setting the stage for a deep MIL revolution \cite{sun2016multiple}. The integration of deep learning transformed MIL, making it a cornerstone of modern weakly supervised learning. Ilse et al. (2018) \cite{ilse2018icml} introduced Attention-based Multiple Instance Learning (ABMIL), a landmark work that used attention to weigh instance contributions, improving both performance and interpretability. This approach became a standard, with refinements like gated attention enhancing focus on critical instances. Shi et al~\cite{shi2020loss} pointed out that the correlation between attention scores and bag label predictions in ABMIL is not strong. Consequently, they proposed to learn the attention scores based on label prediction loss. 
While these attention-based MIL methods \cite{ilse2018icml,shi2020loss,tang2024multi} have advanced the field by adaptively weighting instances, their limitations persist: Firstly, instance redundancy—noisy or irrelevant instances often dilute attention weights, hindering robust classification; (2) Global sparsity—soft attention mechanisms lack explicit mechanisms to discard entire cohorts of non-informative instances.

Compared to approaches utilizing soft weights, many studies prefer to select important instances~\cite{chen2006pami,sofiiuk2019adaptis,li2021deep,zhang2022sparse}, thereby eliminating the influence of unimportant ones. 
The earliest instance selection method is indeed MILES \cite{chen2006pami}. It use $\ell_1$ constraint to select instances with non-zero classification hyperplane parameters, which optimizes the objective function of SVM while considering the $\ell_1$ norm constraint, thus selecting instances that have a significant impact on the classification hyperplane. Similarly, Zhang et al.~\cite{zhang2022sparse} proposed a sparse MIL to select import instances to detect elderly people falling.
However, these methods prefer to select instances within a bag one by one \cite{chen2006pami,zhang2022sparse}, which is instance-level sparsity. For WSI, the number of instances in a bag is very large, the instance-level sparsity fails to consider the local clustering structure of instances. Therefore, it can be inefficient in such a high-dimensional situations.
Subsequently, Li et al.~\cite{li2021deep} perform instance selection via Gumbel softmax and then predict bag label by only use the selected high-scored instances. In their approach, the highest-scoring instances can directly determine the bag’s label. However, in many medical scenarios, this process is influenced by multiple instances of different types collectively determining the bag’s label. For instance, when distinguishing between metastatic and primary mucinous ovarian cancer, it is necessary to identify multiple distinct ROI features corresponding to each type before making a classification decision \cite{dundr2021primary,leen2012pathology}. Moreover, in these existing instance selection frameworks, accurately obtaining the instance-level scores requires iterative computations, leading to relatively high computational costs. 
Zhang et al.~\cite{zhang2022dtfd} propose another attempt for instance selection, wherein instances from a slide are randomly partitioned to form a subset. These subsets are treated as pseudo-bags, which automatically inherit the labels of their parent bags. Subsequently, traditional ABMIL is applied to distill these augmented pseudo-bags, yielding more deterministic pseudo-bag features. These features then proceed to the second stage of ABMIL to obtain improved representations of negative bags. However, directly assigning the parent bag's label to the pseudo-bags in this process introduces additional noise, which can negatively impact the distillation process. 

In this work, we introduce a surprisingly simple yet effective method, Cluster-level Sparse MIL (csMIL), a novel framework that hierarchically integrates global-local instance clustering, within-cluster attention, and cluster-level sparsity induction. Specifically, our approach involves first conducting global clustering on the instances within all the bags to obtain $K$ global cluster centers. Note that the $K$ centers are applicable to all bags. That is why we call them ``global centers''. The number of clusters, $K$, is a hyperparameter that will be selected based on the final classification performance of the bags. Subsequently, using these $K$ global centers as anchors, we perform local clustering within each bag, assigning global $K$-cluster labels to the instances within the bag; this process is referred to as global-local clustering. In the second step, attention scores are calculated for all instances within each cluster in a bag, thereby implementing cluster-level ABMIL. Finally, we assign a corresponding $K$-dimensional cluster weight to each of the $K$ clusters and apply sparse regularization to the cluster weights within the classification loss, which is known as cluster-level sparsification. This cluster sparsification indicates the intention to discard all instances in the bag whose cluster weights are zero, while retaining instances with non-zero weights, thus achieving instance selection at the cluster level.

Our proposed approach offers a systematic framework for cluster-level instance selection without relying on label supervision, which is particularly valuable in computational pathology, where the costs of annotation can be prohibitively high.
It rests on two pillars:
1) Cluster-Attentive Embedding: For each bag, instances are partitioned into $K$ clusters via geometric or semantic similarity. Within each cluster, a local attention mechanism aggregates instances into cluster-specific prototypes, decoupling intra-cluster relevance from inter-cluster importance.  
2) Sparse Cluster Selection: A learnable sparse weight vector, regularized by $\ell_1$-norm, selectively emphasizes or discards entire clusters during bag-level aggregation. This induces global sparsity over clusters rather than instances, enhancing robustness to noisy subgroups.  

Our contributions are as following:
\begin{itemize}
    \item Methodological: A unified clustering-attention-sparsity framework that extends classical MIL pooling operators (e.g., max/mean) and attention-based MIL \cite{ilse2018icml}.
    \item Clinical Utility: Cluster-level sparsity maps pinpoint diagnostically critical regions (validated by pathologists), bridging MIL and interpretable biomarker discovery. 
    \item Theoretical: Recovery bounds showing that $O(s \log K)$ bags suffice to identify $s$ relevant clusters, aligning MIL with compressed sensing principles. 
    \item Empirical: State-of-the-art performance on two histopathology benchmarks CAMELYON16 and TCGA-NSCLC.
\end{itemize}
By unifying sparse learning and structured attention, csMIL advances MIL beyond instance-level heuristics, offering a rigorous, scalable, and interpretable solution for high-stakes applications in computational pathology and beyond.

\section{Related works}

Multiple Instance Learning (MIL) is a weakly supervised learning paradigm designed to handle ambiguously labeled data. Traditional supervised learning requires precise instance-level labels, but in many real-world problems, labels are only available at a higher abstraction level (e.g., a ``bag'' of instances).
The motivation of MIL can be summarized as follows: 1) weak supervision, annotating individual instances (e.g., tumor regions in medical images) is time-consuming, but bag-level labels (e.g., patient diagnoses) are easier to obtain \cite{campanella2019clinical}, 2) label inherent ambiguity. That is, the relationship between instances and labels may be unclear. For instance, in drug discovery, a molecule (bag) is active if at least one conformation (instance) is effective \cite{dietterich1997solving}. For another example, in a review (bag) labeled as ``negative'', only 2 sentences contain explicit complaints (positive instances), while the other 10 sentences describe product functions (negative instances).  Although a standard MIL works by assume that a bag is positive if at least instance is positive \cite{dietterich1997solving}, however, people find that bag labels also depend on instance interactions. For example, majority voting \cite{ramon2000multi} or presence/absence of key instances \cite{liu2012key,carbonneau2018multiple}. The threshold-based assumptions is also popular, it claim that a bag is positive if a minimum number of instances are positive \cite{kotzias2015group}. 
3) Another motivation of MIL is ``cost efficiency''. MIL Leverages coarse labels to train models for fine-grained tasks, such as tumor localization in histopathology\cite{ilse2018icml}.

To solve the MIL problem, people try to extend some traditional machine learning methods, such as mi-SVM/MI-SVM \cite{Andrews2002MISVM}, which extend SVMs to MIL by iteratively refining instance labels, and EM-DD \cite{zhang2001dd}, which combines expectation-maximization with diverse density to identify key instances. Recently, some deep-learning based methods have been proposed ito adapt the MIL problem. For example, the attention-based MIL \cite{ilse2018icml} learn instance-specific weights to emphasize critical instances with attention mechanism. Zaheer et al.\cite{zaheer2017deep} propose a permutation-invariant networks to handle variable-sized bags. Wu et al. \cite{wu2015deep} propose to aggregate instance features into a fixed-size bag representation using Max/Mean pooling. In \cite{shao2021transmil}, a Transformer-based MIL has been proposed. It leverages self-attention for modeling instance interactions.

MIL is widely used in domains with weakly labeled data, such as drug discovery \cite{dietterich1997solving}, image classification \cite{li2017pami}, weakly supervised object detection \cite{cinbis2016weakly}, video analysis \cite{gu2008multi,shao2023video}, document classification (bag = document, instances = sentences or paragraphs) \cite{kotzias2015group,he2009text}, disease classification using MRI/CT scans with patient-level labels \cite{tong2014multiple} and whole-slide histopathology images \cite{courtiol2018classification,campanella2019clinical,yao2020whole,li2021dual,zhu2022murcl}. 

It is worth mentioning that whole-slide images (WSI) analysis is currently one of the most important application scenarios for multi-instance learning \cite{wang2024advances}, and it is also the focus of this paper. MIL treats WSIs as ``bags'' of patches (instances), with only bag-level labels available during training. A cornerstone of MIL, Attention-Based MIL (ABMIL) \cite{ilse2018icml}, remains widely used for its interpretable attention mechanisms. A 2024 survey by Liu et al. (arXiv:2408.09476) highlights ABMIL’s role in cancer classification, detection, and molecular marker prediction, while noting its limitations in instance discrimination. To address these, Cai et al. (2024) \cite{cai2024rethinking} proposed Attribute-Driven MIL (AttriMIL), which uses attribute scoring and spatial constraints to model instance correlations. Similarly, Zhang et al. (2024) \cite{zhang2024aem} introduced Attention Entropy Maximization (AEM), a regularization method that prevents overfitting by distributing attention across informative WSI regions, enhancing model generalization.


Shifting focus to instance-level classification, Qu et al. (2023) proposed Key Patches Are All You Need (KPAN) \cite{qu2024rethinking}, a framework that emphasizes a strong patch-level classifier using contrastive and prototype learning. KPAN outperforms complex MIL architectures on CAMELYON16 by avoiding noisy pseudo-labels and misclassified patches. Similarly, Liu et al. (2024, arXiv:2408.09449) introduced FocusMIL \cite{liu2024attention}, which uses max-pooling and variational inference to prioritize tumor morphology over spurious correlations like staining conditions, improving interpretability and recognition of challenging instances. 

For fine-grained classification, Jin et al. (2024) developed Hierarchical MIL (HMIL) \cite{10810475}, which aligns label correlations at instance and bag levels to capture subtle morphological variations critical for precision oncology. Meanwhile, Li et al. (2025) proposed Channel Attention-Based MIL (CAMIL) \cite{mao2025camil}, integrating channel attention with transformers via a Multi-scale Channel Attention Block to enhance spatial and channel feature representation, outperforming TransMIL. Addressing tumor heterogeneity, Wang et al. (2023) introduced Multiplex-Detection-Based MIL (MDMIL) \cite{wang2023targeting}, which uses multiplex detection to identify critical instances in heterogeneous tumor regions, improving performance on datasets like TCGA Lung Cancer.

To enhance robustness, Chen et al. (2024) developed Causal MIL (CaMIL) \cite{chen2024camil}, a framework that uses interventional probability to distinguish causal from spurious associations, such as stain color, improving accuracy across diverse WSIs. Computational efficiency was tackled by Keshvarikhojasteh et al. (2024) through random patch sampling \cite{keshvarikhojasteh2024multiple}, achieving performance gains of 1.7\% on CAMELYON16 and 3.7\% on TUPAC16 with reduced computational cost, though interpretability effects varied by dataset.

These advancements address key MIL challenges: KPAN and AttriMIL mitigate noisy pseudo-labels, AEM and FocusMIL prevent overfitting, MDMIL and CaMIL tackle tumor heterogeneity, and random sampling reduces computational demands. Interpretability is enhanced through reliable heatmaps and tumor localization, fostering clinical trust. Future directions include developing robust models for diverse diseases, optimizing for real-time inference, refining attention mechanisms for actionable insights, and integrating MIL with multi-omics data for comprehensive cancer profiling. Collectively, these innovations position MIL as a transformative tool for precision oncology.




MIL bridges the gap between weak supervision and robust model training, enabling applications where fine-grained labels are impractical. Its theoretical diversity and adaptability to domains like healthcare and computer vision make it a vital area in machine learning research. As deep learning advances, MIL continues to evolve, offering scalable solutions for real-world ambiguity.
Our method can be seen as a hybrid of attention-based MIL and sparse coding. It generalizes Ilse et al.’s model by introducing a clustered intermediate representation and connects to sparse MIL by enforcing cluster-level sparsity, potentially improving robustness to noisy or irrelevant instances. Sparse MIL Variants:  We use of $\ell_1$ regularization on $\beta$ aligns with sparse feature selection techniques in MIL (e.g., sparse SVM-based MIL). The key difference is that sparsity here operates at the cluster level rather than the instance level, offering a coarser granularity of selection.

\section{Model}
\subsection{Attention-based Multiple Instance Learning}
Suppose that we have a whole-slide image dataset $\left\{ \left( X_i, y_i\right)\right\}_{i=1}^M$, for each slide $X_i = \{\mathbf{x}_i\}_{i=1}^{N_i}$ contains $N_i$ patches (or instances). we consider neural networks $g_{\theta}(\cdot)$ with parameters $\theta$ that extracts the $j$-th instance's feature as a low-dimensional embedding $\mathbf{h}_j$. Then we get the low-dimensional representation $H_i = \{\mathbf{h}_i\}_{i=1}^{N_i}$ for slide $X_i$. Traditional MIL works as follows,
\begin{equation}
    \mathbf{z}=\sum_{n=1}^{N_i} \alpha_n\mathbf{h}_n
\end{equation}
that is, we aggregate all instances in bag $X_i$ with a weight of $\alpha_n$ to obtain a bag representation $\mathbf{z}$. The weight $\alpha_n$ could be a function of $H_i$,
\begin{equation}
    \alpha_n = f(H_i).
\end{equation}
There are numerous choices for the function $f$. It could be a constant, such as $f=\frac{1}{N}$,  or it might take the form of $f=\max(\cdot)$. Additionally, $f$ can be learned using methods like  $L_1$  graph \cite{5357420}, $\epsilon$-graph \cite{zhou2009icml}, and attention mechanisms \cite{ilse2018icml}.

Among these, the attention-based multiple-instance learning (MIL) method \cite{ilse2018icml} is particularly popular. In this approach, $f$ is defined as follows:
\begin{equation}
    \alpha_n = \frac{\exp( \mathbf{w}^{\top}\text{tanh}({\mathbf{V}\mathbf{h}_n}^{\top}))}{\sum_j \exp( \mathbf{w}^{\top}\text{tanh}({\mathbf{V}\mathbf{h}_j}^{\top}))}.
\end{equation}

\subsection{Kick negative instances out }
\begin{figure*}[ht]
    \centering
    \includegraphics[width=0.8\textwidth]{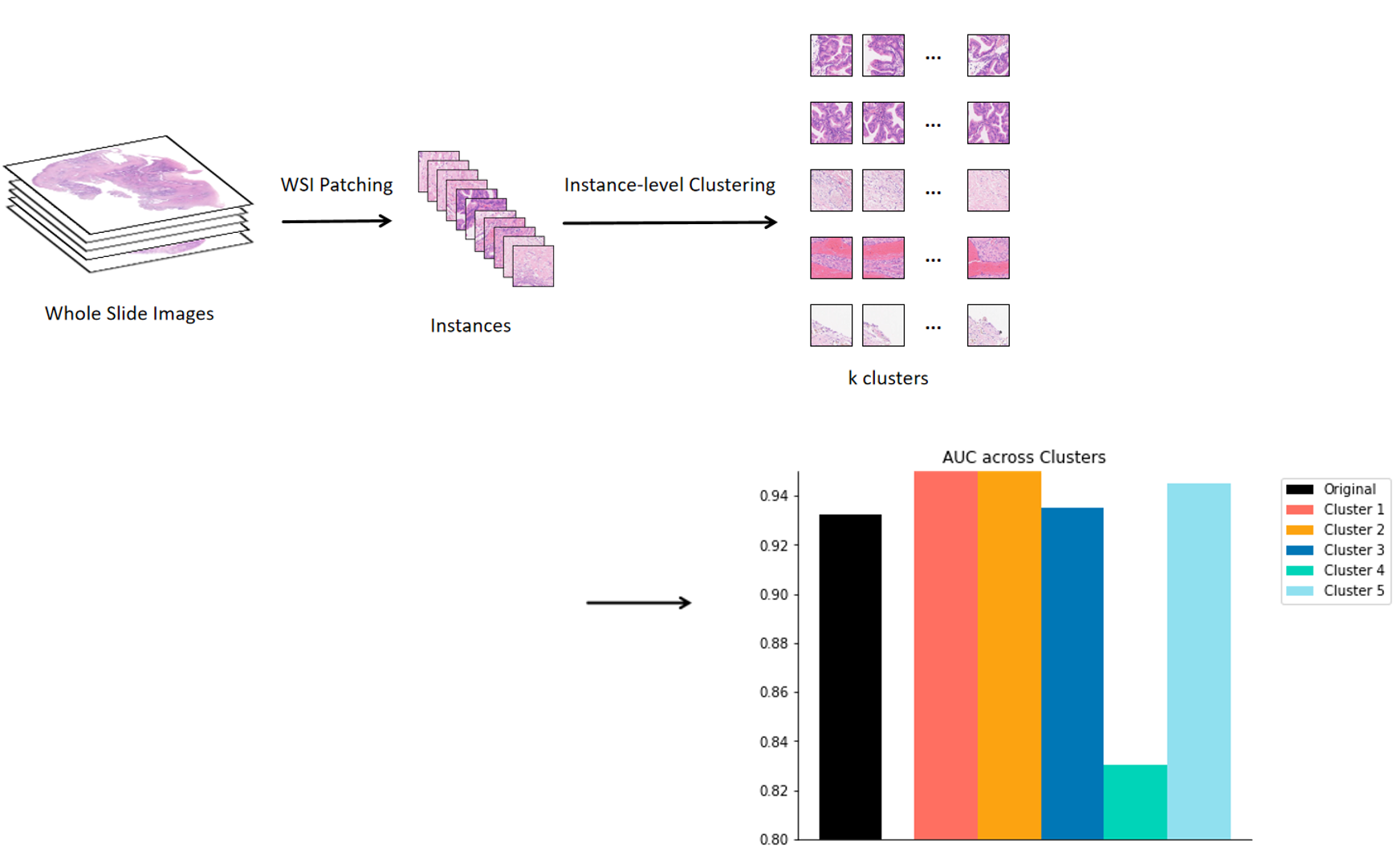}
    \caption{Under the framework of ABMIL, all samples are sliced to obtain instances, and these instances are clustered into five groups. Finally, a performance comparison is conducted by attempting to remove the instances of one cluster at a time from the five clusters.}
    \label{fig:motivation}
\end{figure*}

As discussed previously, early studies in instance learning tend to construct bag representations by averaging all instances. However, in multi-instance learning scenarios, not all instances contribute positively to the performance of bag classification, making simple average pooling suboptimal. We define instances that positively influence bag classification as positive instances, while others are considered negative instances. Existing research has primarily focused on attention-based mechanisms designed to suppress negative instances while enhancing the significance of critical positive ones. This leads us to our hypothesis: \textit{Could the complete elimination of negative instances from the pooling process fundamentally mitigate their adverse effects?}
Unfortunately, the core challenge lies in the lack of prior knowledge regarding instance labels.

We propose a clustering-driven solution based on the assumption that positive and negative instances reside in distinct clusters. Our approach involves systematically evaluating classification performance following the exclusion of each cluster. For instance, given $K$ clusters of instances, we iteratively remove one cluster at a time and monitor changes in classification performance. For example, if the exclusion of cluster $c_i$ results in a performance drop, it is likely that to claim that $c_i$ is a positive cluster. Conversely, an improvement in performance suggests that $c_i$ predominantly consists of negative instances. 

In fact, the above conjecture can be easily verified. Figure \ref{fig:motivation} illustrates this workflow and provides experimental validation on a whole-slide image dataset. We first clustered all instances into 5 clusters. Then, we selected ABMIL~\cite{ilse2018icml} as the baseline model for multiple experiments. We removed one of the 5 clusters at a time, measured the performance of the slide classification, and ensured that all other settings remained identical across the experiments. The final results are presented in the bar plot in Figure \ref{fig:motivation}. The black bar represents the classification accuracy of ABMIL when no instances are discarded, while the subsequent five bars represent the classification accuracy of the ABMIL model after removing clusters 1 through 5, respectively. We observed that removing cluster 4 led to a significant decline in classification performance, whereas removing clusters 1, 2, and 5 actually improved the performance.

It is noteworthy that while we categorize all instances into positive and negative instances, the number of clusters $K$ in clustering does not necessarily have to equal 2; in many scenarios, it is necessary to set $K > 2$. For instance, when diagnosing mucinous ovarian cancer using whole slide imaging (WSI) data, although the aim is to classify metastatic ovarian cancer and primary ovarian cancer, the features supporting these classifications are diverse. For example, features indicative of primary ovarian cancer may include smooth capsule, evenly distributed cystic and solid areas, and areas of mucinous cystadenoma, among others. In contrast, features supporting metastatic ovarian cancer may consist of signet ring cells, involvement of surface and superficial cortex, and infiltrative (destructive) invasion \cite{leen2012pathology}. Thus, each category is determined by a variety of slide features rather than solely by clustering characteristics, suggesting that $ K $ should be set to a value related to the total number of slide features.

The results in Figure \ref{fig:motivation} indicate that the classification outcomes either improve or decline when one of the five clusters is removed, thereby validating our hypothesis. Furthermore, if two clusters are eliminated from the five, what impact would this have on the classification results? In this scenario, we would need to conduct $\tbinom{5}{2} = 10$ experiments to ascertain the outcome. Similarly, to verify the removal of three clusters, $\tbinom{5}{3} = 10$ experiments would be necessary. If the number of clusters is set larger, identifying the corresponding clusters for positive and negative instances would require even more experiments. Is there a more efficient method that could allow us to determine all results in a single attempt? This is precisely what we will introduce in the next section.

\subsection{The proposed method:Cluster-level sparse multi-instance learning}

As previously discussed, we have $M$ slides (bags), with each slide containing $N_i$ patches (instances). Consequently, the total number of instances is given by $\sum_{i = 1}^M N_i=N$.
The proposed cluster-level sparse multi-instance learning method operates as follows. 
First, we group all the $N$ training patches into $K$ clusters using $K$-means, $\left\{ \mathbf{h}_i\right \}_{i=1}^{C_1}$, \ldots,$\left\{ \mathbf{h}_i\right \}_{i=1}^{C_k}$, \ldots, $\left\{ \mathbf{h}_i\right \}_{i=1}^{C_K}$, that is $C_1+\cdots+C_K=N$, then we can obtain $K$ global cluster centers $\{ \bar{\mathbf{h}}_1, \ldots,\bar{\mathbf{h}}_K\}$ and a low-dimensional representation $H_i = \{\mathbf{h}_1,\ldots,\mathbf{h}_{N_i}\}$ for slide $X_i$. We call this process global clustering.

Due to the variations between bags, inconsistencies may arise in the global clustering results on individual bags. Directly clustering each slide separately, even with the same number of clusters, can lead to misalignment of the cluster centers across different slides. 
Therefore, we employ the aforementioned global-local combined clustering approach for data preprocessing. 
That is, we perform local clustering within each slide $H_i = \{\mathbf{h}_j\}_{j=1}^{N_i}$ by assigning $K$ fixed centers $\{ \bar{\mathbf{h}}_1,\ldots,\bar{\mathbf{h}}_K\}$, and obtain at most $K$ clusters $\left\{ \mathbf{h}_j\right \}_{j=1}^{C^i_1}$, \ldots,$\left\{ \mathbf{h}_j\right \}_{j=1}^{C^i_k}$, \ldots, $\left\{ \mathbf{h}_j\right \}_{j=1}^{C^i_K}$ in slide $X_i$, where $C^i_k$ is the number of instances belonging to cluster $k$ in bag $i$; that is $C^i_1+\cdots+C^i_K=N_i$.

Following these two rounds of global-local clustering, all instances within each slide (bag) can be assigned to $K$ cluster labels based on $K$ specified global-shared cluster centers. It is important to note that the final number of clusters obtained may be less than $K$, as some slides may not contain instances that exist globally. For example, during global clustering, a certain cluster may represent blood vessels, but other slides may not necessarily contain any blood vessels.

With the global cluster labels for all the instances, we propose to run a traditional attention-based MIL on each cluster $\left\{ \mathbf{h}_j\right \}_{j=1}^{C^i_k}$,
\begin{equation*}
    \mathbf{z}_k^i=\sum_{i=1}^{C^i_k} \alpha_i^k \mathbf{h}_i.
\end{equation*}
For the $i$-th bag, we obtain representations $Z_i:=\{\mathbf{z}_1^i,\ldots,\mathbf{z}_k^i,\ldots,\mathbf{z}_K^i\}$, $i=1,\ldots,M$. For each of its clusters, we aggregate bag representation $\mathbf{z}^i$ for slide $X_i$ as  
\begin{equation*}\label{eq:clusterWeighted}
    \mathbf{z}^i=\sum_{k=1}^{K} \beta_k\mathbf{z}_k^i ,
\end{equation*}
where $\beta_k$ represents the corresponding weight of cluster $k$ in slide $i$.
It is important to note that $\boldsymbol{\beta}=(\beta_1,\cdots, \beta_K)$ is learnable parameters, and its significance differs from that of the previously mentioned $\boldsymbol{\alpha}=(\alpha^1,\ldots, \alpha^{C_k^i})$.

With the bag's representation $\mathbf{z}^i$, we can calculate the classification loss over all the bags as
\begin{equation}\label{eq:loss}
    Loss = \sum_{i=1}^M CE(g(\mathbf{z}^i),Y^i)+\gamma \|\boldsymbol{\beta} \|_1
\end{equation}
where $M$ is the number of bags, $g(\cdot)$ is a linear function that link $\mathbf{z}^i$ to a probability simplex, and $CE$ represents the prediction cross-entropy loss for the slide. The second term in the loss function imposes a $\ell_1$ norm constraint on the parameter $\boldsymbol{\beta}$, requiring a sparse structure. Therefore, if the $k$-th component of the estimated $\boldsymbol{\beta}$ vector is close to 0 or equal to 0, this indicates that the $k$-th cluster has zero importance in the aggregation process illustrated in Equation (\ref{eq:loss}), 
implying that our model discards cluster $k$. Similarly, if multiple components of $\boldsymbol{\beta}$ are equal to 0, it indicates that there are multiple clusters corresponding to instances that should also be discarded, as they contribute nothing to the final classification task.

\section{Theoretical Analysis of the sparseMIL}
This framework bridges instance selection and representation learning in MIL, offering theoretical soundness comparable to group lasso while maintaining the expressiveness of attention-based deep MILs.
Nonzero $\beta_k$ directly map to clinically meaningful clusters (e.g., tumor subregions in WSI). 
\subsection{Relation to Existing MIL Paradigms}

The proposed method has a close relationship with three fundamental concepts,
\begin{enumerate}
    \item ABMIL \cite{ilse2018icml}:
The proposed approach builds upon the instance-level attention mechanism, as defined in Equation (\ref{eq:clusterWeighted}). This extension results in a hierarchical structure, where the attention mechanism first operates at the local level within individual clusters and then aggregates globally. This hierarchical design enables a more nuanced and context - aware processing of the data within the multiple instance framework.
   \item Prototype Learning \cite{li2021adaptive}:
In the proposed method, the cluster - specific representations $\mathbf{z}_k^i$ serve as learned prototypes. This concept is analogous to that in prototype networks, where individual instances are compressed into representative vectors. These prototypes capture the essential characteristics of the instances within their respective clusters, facilitating a more efficient and meaningful representation of the multiple instance data.
\item Instance-level sparse MIL \cite{zhang2022sparse}. The use of $\ell_1$-regularization in the proposed method induces instance-level sparsity. This is functionally equivalent to group sparsity methods, which are designed to eliminate irrelevant subsets of instances. By promoting sparsity at the cluster level, the method can effectively filter out noise and focus on the most relevant information within the multiple instance data, enhancing both the interpretability and performance of the model. 
\end{enumerate}

\subsection{Computational Complexity}
The method introduces overhead from two components: 
\begin{enumerate}
    \item Clustering, the complexity is $O(M \cdot \max_i N_i \cdot K \cdot d)$ for $K$ clusters and $d$-dimensional embeddings. But it is offline.
    \item Sparse Optimization, $\ell_1$ regularization requires proximal gradient methods. The iteration cost is $O(MK^2 + MKd)$ vs. standard attention MIL's $O(MN^2 + MNd)$. In our scenario, $K \ll N$, the method reduces cost significantly. 
\end{enumerate}

\subsection{Theoretical Guarantees}
\begin{theorem}\label{theorem:sparseRecovery}
Sparse Recovery \cite{candes2008enhancing}: If true discriminative prototypes reside in $s \leq K$ clusters, $\ell_1$ regularization achieves exact support recovery with probability $1-\delta$ when $M \geq C s \log(K/\delta)$, where $C$ is a constant.
\end{theorem}

Theorem \ref{theorem:sparseRecovery} that applying $\ell_1$  regularization on the prototype weights $\boldsymbol{\beta}$ to induce sparsity, aiming to select a small number of relevant clusters (prototype) out of $K$ total clusters. The claim is that if the true model has $s$ significant prototypes (with $s \leq K$), then with high probability $1-\delta$ , the correct clusters can be recovered when the number of bags $M$ satisfies $M \geq C s \log(K/\delta)$. 
The constant $C$ typically depends on properties of the measurement matrix, such as the restricted isometry property (RIP) or coherence.




Under such an approximation, the design matrix would consist of the cluster-aggregated features $\mathbf{z}_i$ for each bag, and the labels $Y_i$. The $\ell_1$-regularized loss corresponds to the Lasso problem. In Lasso theory, the sample complexity for exact recovery is known to be $M \geq C s \log(K/\delta)$, where $C$ depends on the restricted eigenvalue (RE) condition of the design matrix.

But in the MIL setup, the design matrix isn't directly observed; instead, the features $\mathbf{z}_i$ are learned through the network. This complicates things because the features are data-dependent and learned. However, if we assume that the learned features satisfy certain properties (like the RE condition), then the Lasso recovery bounds could apply. The constant $C$ would then depend on properties like the minimum eigenvalue of the covariance matrix of the relevant clusters, the coherence between clusters, and the noise level in the labels.

Additionally, the $\log (K/\delta)$ term arises from the probabilistic nature of the recovery guarantee. The logarithmic dependence on $K$ is due to the union bound over all possible subsets of clusters, and $\delta$ controls the failure probability.

To summarize, the sparse recovery bound suggests that with sufficiently many bags $M$, the $\ell_1$-regularized objective can correctly identify the non-zero clusters with high probability. The constant $C$ encapsulates the difficulty of the problem, depending on the data distribution and the properties of the cluster features. However, deriving the exact form of 
$C$ would require a more rigorous analysis, possibly under specific assumptions about the data generation process and the feature extractor.

\section{Results}
\begin{figure*}[ht]
    \centering
    \subfigure[Camelyon16]{       \includegraphics[width=0.45\textwidth]{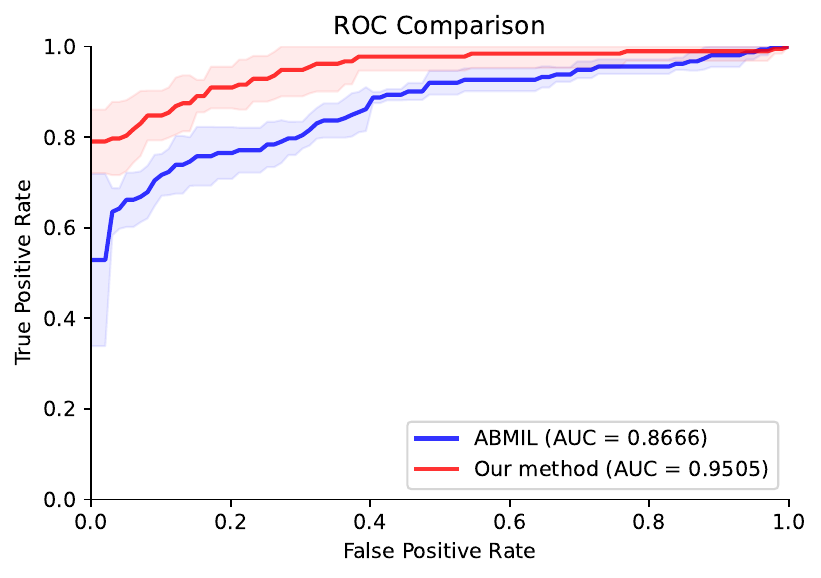}
    }
    \subfigure[TCGA-NSCLC]{        \includegraphics[width=0.45\textwidth]{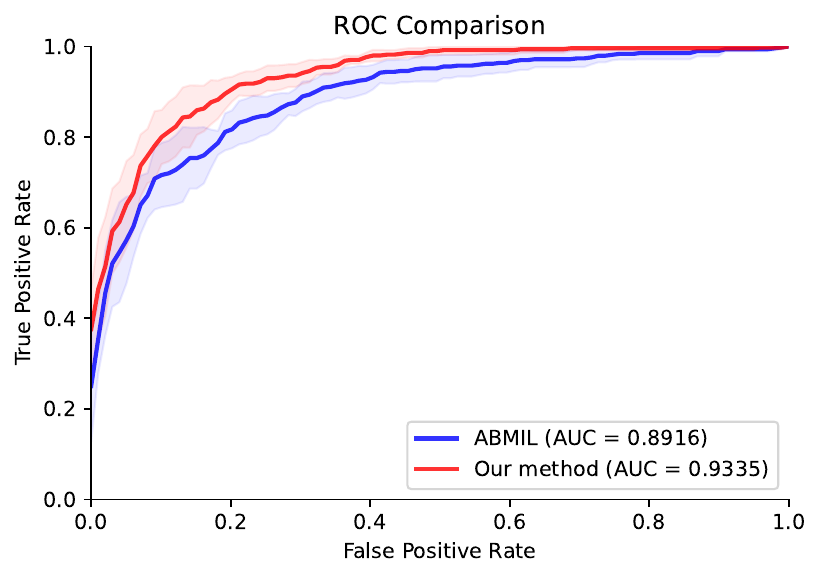}
    }
    \caption{Comparisons with the original ABMIL model under the setting of $K=3$.}
    \label{fig:roc}
\end{figure*}

\subsection{Datasets}
The CAMELYON-16 dataset, a well-known publicly available medical image dataset, is predominantly employed for assessing algorithms aimed at detecting lymph node metastases in breast cancer \cite{bejnordi2017diagnostic}. Comprising 400 whole slide images (WSIs), this dataset was developed by the Radboud University Medical Center and the University Medical Center Utrecht, both located in the Netherlands. It was utilized during an international competition in 2016. The dataset encompasses hundreds of high-resolution WSIs obtained from lymph node sections of breast cancer patients. These sections are stained to distinctly visualize the presence or absence of cancer cells, thereby facilitating the training and evaluation of automated cancer detection systems.

The Cancer Genome Atlas (TCGA) lung cancer whole-slide image (WSI) dataset \cite{cancer2012comprehensive} is another significant dataset. It consists of 1046 WSIs representing two cancer subtypes: lung adenocarcinoma (LUAD) and lung squamous cell carcinoma (LUSC).

Regarding the pre-processing step, non-overlapping patches of $256 \times 256$ pixels at a $20\times$ magnification were extracted from each WSI in both datasets. For sample splitting, to conduct five-fold cross-validation, we partitioned these two datasets into training and test sets at the patient level with a ratio of 4:1.

\subsection{Baselines}
We select the following baseline methods for performance comparisons:
\begin{itemize}
    \item ABMIL~\cite{ilse2018icml}, Attention-based Deep Multiple Instance Learning.
    \item MinMax~\cite{courtiol2018classification}, Classification and disease localization in histopathology using only global labels: A weakly-supervised approach.
    \item RNNMIL~\cite{campanella2019clinical}, Clinical-grade computational pathology using weakly supervised deep learning on whole slide images.
    \item SetTransformer~\cite{lee2019set}, Set transformer: A framework for attention-based permutation-invariant neural networks.
    \item DeepAttnMIL~\cite{yao2020whole}, Whole slide images based cancer survival prediction using attention guided deep multiple instance learning networks.
    \item DSMIL~\cite{li2021dual}, Dual-stream multiple instance learning network for whole slide image classification with self-supervised contrastive learning.
    \item CLAM-MB~\cite{lu2021data}, Data-efficient and weakly supervised computational pathology on whole-slide images
    \item DTFD-MIL~\cite{zhang2022dtfd}, Dtfd-mil: Double-tier feature distillation multiple instance learning for histopathology whole slide image classification.
\end{itemize}

\subsection{Overall Performance}
\begin{table}[ht]
    \centering
    \setlength{\belowcaptionskip}{10pt} 
    \caption{The performance of the proposed method is compared with the baseline algorithms using five-fold cross-validation on the CAMELYON-16 dataset.}
    \begin{tabular}{@{}lccc@{}}
        \toprule
        Method & Acc & F1 & AUC \\ \midrule
        ABMIL~\cite{ilse2018icml} & 0.863& 0.797 & 0.891 \\
        MinMax~\cite{courtiol2018classification} & 0.873 & 0.814 & 0.904\\
        RNNMIL~\cite{campanella2019clinical} & 0.848 & 0.762 & 0.861 \\
        SetTransformer~\cite{lee2019set} & 0.882 & 0.824 & 0.908 \\
        DeepAttnMIL~\cite{yao2020whole}  & 0.887 & 0.831 & 0.927 \\
        DSMIL~\cite{li2021dual} & 0.876 & 0.820 & 0.907 \\
        CLAM-MB~\cite{lu2021data} & 0.862 & 0.809 & 0.906 \\
        CLAM-SB~\cite{lu2021data} & 0.858 & 0.806 & 0.906 \\
        DTFD-MIL~\cite{zhang2022dtfd} & 0.898 & 0.872 & 0.946 \\
        Our method  & \textbf{0.905} & \textbf{0.876} & \textbf{0.951} \\
        \bottomrule
    \end{tabular}
    \label{tab:camelyon16_results}
\end{table}
Table \ref{tab:camelyon16_results} illustrates the performance of the baseline methods and our method on CAMELYON-16 dataset. The evaluation metrics considered are Accuracy (Acc), F1-score, and Area Under the Curve (AUC).
We see that the proposed method achieved the highest accuracy (0.905), F1 score (0.876), and AUC (0.951) among all methods compared. The next best was DTFD-MIL with an accuracy of 0.898, followed by DeepAttnMIL (0.887) and SetTransformer (0.882).  RNNMIL was the lowest performer across all metrics, with an accuracy, F1, and AUC at 0.848, 0.762, and 0.861 respectively.

 \begin{table}[ht]
    \centering
    \setlength{\belowcaptionskip}{10pt} 
    \caption{The performance of the proposed method is compared with the original ABMIL and other baselines using five-fold cross-validation on the TCGA-NSCLC dataset.}
    \begin{tabular}{@{}lccc@{}}
        \toprule
        Method & Acc & F1 & AUC \\ \midrule
        ABMIL~\cite{ilse2018icml} & 0.783 & 0.767 & 0.866 \\
        MinMax~\cite{courtiol2018classification} & 0.844 & 0.834 & 0.918 \\
        RNNMIL~\cite{campanella2019clinical} & 0.736 & 0.739 & 0.818 \\
        SetTransformer~\cite{lee2019set} & 0.844 & 0.834 & 0.918 \\
        DeepAttnMIL~\cite{yao2020whole} & 0.753 & 0.747 & 0.814 \\
        DSMIL~\cite{li2021dual} & 0.846 & 0.848 & 0.914 \\
        CLAM-MB~\cite{lu2021data} & 0.832 & 0.810 & 0.914 \\
        CLAM-SB~\cite{lu2021data} & 0.837 & 0.816 & 0.919 \\
        DTFD-MIL~\cite{zhang2022dtfd} & 0.845 & 0.833 & 0.922 \\
        Our method  & \textbf{0.854} & \textbf{0.855} & \textbf{0.933} \\
        \bottomrule
    \end{tabular}
    \label{tab:tcga_results}
\end{table}

Table \ref{tab:tcga_results} summarizes the comparisons between the baseline methods and our approach on the TCGA-NSCLC dataset. The results reveal that our method again achieved the highest accuracy (0.854), F1 score (0.855), and AUC (0.933). DSMIL, DTFD-MIL, and MinMax exhibited relatively close accuracy values, with 0.846, 0.845, and 0.844, respectively. Compared with the accuracy, our method has achieved greater improvements in the AUC and the F1 score compared to other benchmark methods.  
These findings highlight the superior performance of our proposed method compared to existing models when applied to the TCGA-NSCLC dataset.

Our method has demonstrated consistent superiority over the baseline methods on CAMELYON-16 and TCGA-NSCLC datasets. The performance improvements in accuracy, F1 score, and AUC on the two datasets imply that our proposed method is more adept at extracting relevant instances from the image bags, which is crucial for accurate whole-slide image classification. 
The proposed method builds upon the framework of ABMIL, with the key differentiator being the integration of sparse clustering regularization, which serves to filter out redundant instances. Given this fundamental distinction, a comprehensive analysis of the performance disparities between ABMIL and the proposed methodology is imperative.
To this end, Figure \ref{fig:roc} presents a visual comparative assessment of the receiver operating characteristic (ROC) curves for both methods. Panel (a) depicts the results obtained on the Camelyon16 dataset, while panel (b) showcases the findings from the TCGA-NSCLC dataset. A direct comparison of these ROC curves reveals a pronounced performance superiority of the proposed method over ABMIL. This outcome underscores the critical role played by the introduced sparse clustering regularization in augmenting diagnostic accuracy, thereby demonstrating its efficacy in improving the overall performance of the classification system.

\subsection{Choices of number of clusters $K$}
The experiments evaluate the impact of cluster number ($K$) on the performance of our method across the two real-world datasets CAMELYON-16 and TCGA-NSCLC as shown in Table \ref{tab:camelyon16_results} and \ref{tab:tcga_results} respectively.

In the analysis of the CAMELYON-16 dataset, as presented in Table \ref{tab:camelyon16_results}, several key observations can be drawn. First, the optimal number of clusters, denoted as $K=3$, achieves the highest performance across all evaluated metrics, specifically an accuracy of 0.905, an F1 score of 0.876, and an AUC of 0.951. Second, the implementation of close cluster ($K=2$ or $K=5$) yield competitive results; however, the use of relative larger cluster ($K=10$) result in a degradation of AUC. This decline suggests that over-segmentation introduces additional noise into the analysis. Notably, while $K=10$ demonstrates a slight recovery in accuracy (0.892), it exhibits a reduction in AUC (0.931), indicating that fragmented clusters detrimentally affect discriminative power.

Table \ref{tab:tcga_results} outlines the results obtained from the TCGA-NSCLC dataset. Similar trends are observed. First, the optimal configuration of $K=3$ outperforms the alternatives again, achieving an accuracy of 0.854, an F1 score of 0.855, and an AUC of 0.933. Furthermore, the use of larger clusters ($K=5, 10$) is associated with significant drops in performance, with AUC values falling to 0.907 or lower. This decline is likely attributed to the dilution of sparse features within the high-dimensional genomic data. Conversely, the configuration with $K=2$ demonstrates robustness, achieving an accuracy of 0.852 and an AUC of 0.925, suggesting that simpler clustering structures may be adequate for certain genomic subtyping tasks.

Based on the above results, although we found that when $K=3$, our method achieved the best performance on both datasets, the trends of performance changes were not the same under other settings of $K$. This indicates that the selection of $K$ is related to the dataset.

\begin{table}[!htp]
    \centering
    \setlength{\belowcaptionskip}{10pt} 
    \caption{The performance of the our method with different number of cluster settings on the CAMELYON-16 dataset.}
    \begin{tabular}{@{}cccc@{}}
        \toprule
        Number of cluster & Acc & F1 & AUC \\ \midrule
        $K=2$  & 0.894 & 0.865 & 0.942\\
        $K=3$  & \textbf{0.905} & \textbf{0.876} & \textbf{0.951} \\
        $K=5$  & 0.887 & 0.849 & 0.938 \\
        $K=10$  & 0.892 & 0.865 & 0.931 \\
        \bottomrule
    \end{tabular}
    \label{tab:diffClusterCamelyon-16}
\end{table}

For whole-slide image analysis, small $K$ (e.g., $K=2$) captures coarse-grained patterns but risks under-segmenting heterogeneous instances (e.g., tumor subregions in CAMELYON-16). This balances noise suppression with limited representation capacity.  For moderate $K$ (e.g., $K=3$), our method strikes an optimal trade-off. That is, the clusters are sufficiently fine to model intra-bag diversity (e.g., distinct cancer subtypes in TCGA-NSCLC) while avoiding overfitting. The sparse aggregation mechanism effectively prunes irrelevant clusters.  In contrast, if we keep increasing$K$, e.g., $K=5, 10$, the over-segmentation leads to redundant clusters, diluting discriminative features. In TCGA-NSCLC, high-dimensional genomic features exacerbate this issue, as noisy clusters dominate the sparse aggregation.



\begin{table}[ht]
    \centering
    \setlength{\belowcaptionskip}{10pt} 
    \caption{The performance of the our method with different number of cluster settings on the TCGA dataset.
    }
    \begin{tabular}{@{}lccc@{}}
        \toprule
        Number of cluster & Acc & F1 & AUC \\ \midrule
        $K=2$  & 0.852 & 0.852 & 0.925 \\
        $K=3$  & \textbf{0.854} & 
        \textbf{0.855} & \textbf{0.933} \\
        $K=5$  & 0.823 & 0.826 & 0.907 \\
        $K=10$  & 0.824 & 0.828 & 0.900 \\
        \bottomrule
    \end{tabular}
    \label{tab:diffClusterTCGA}
\end{table}










\subsection{Interpretations from the perspective of medicine}
\begin{figure*}[ht]
    \centering
    \subfigure[Original Visualization Results of Sample tumor51.]{
        \includegraphics[width=0.45\textwidth]{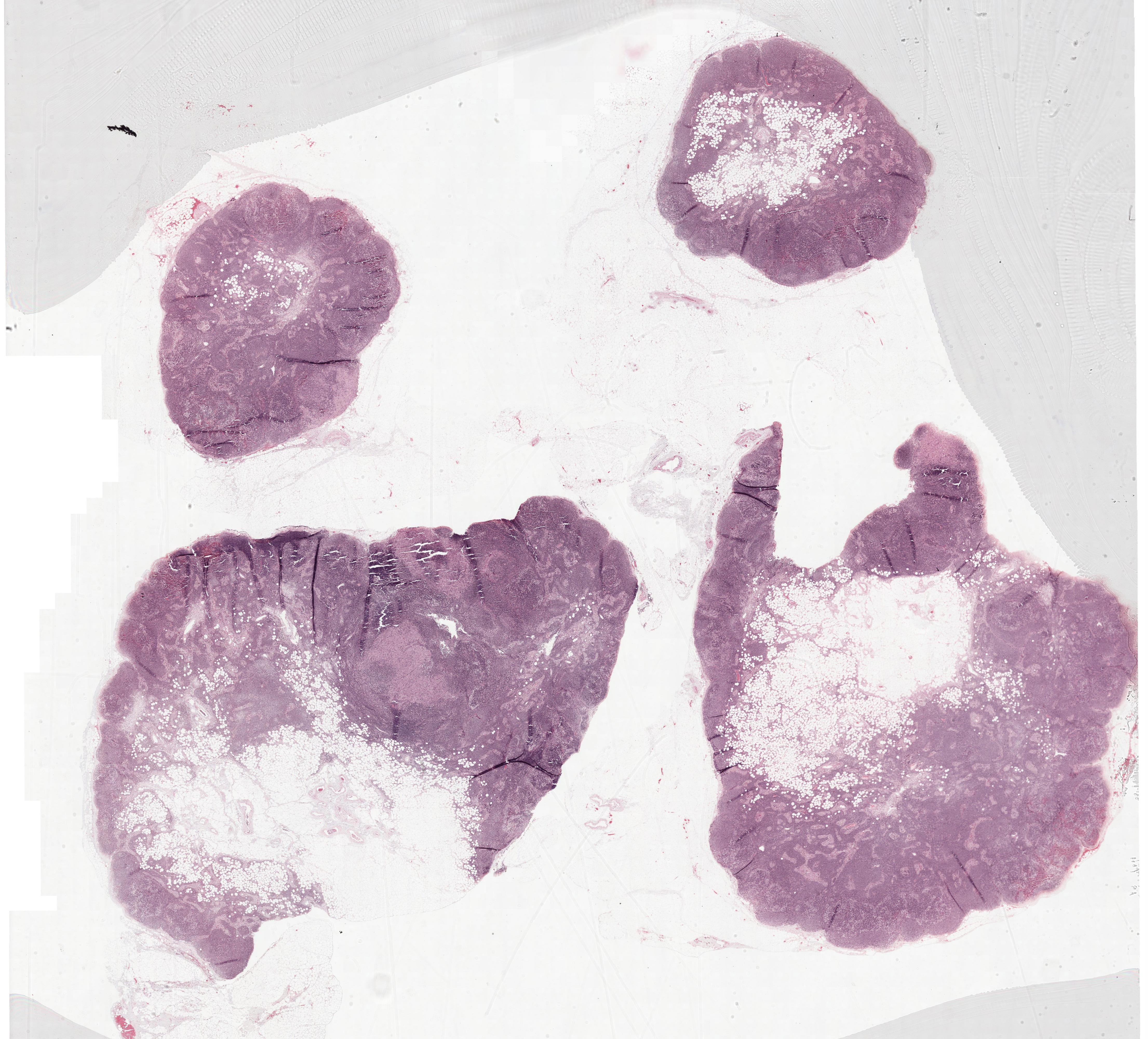}
    }
    \hfill  
    \subfigure[Visualization Results of Sample tumor52 under Two-Class Clustering $\boldsymbol{\beta}=(1.04,-1.59)$. Red: cluster 1, blue: cluster 2.]{
        \includegraphics[width=0.45\textwidth]{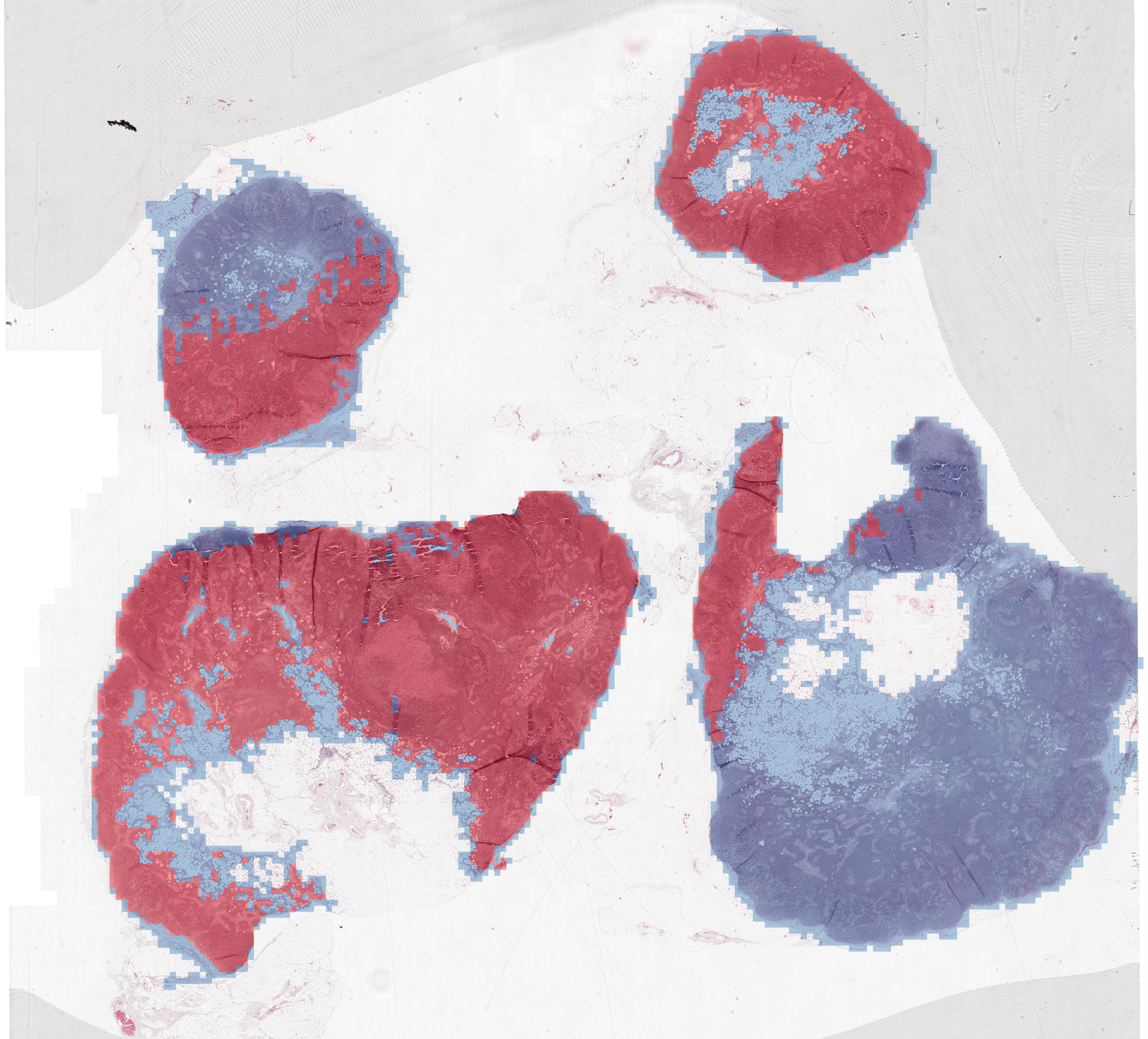}
    }
    \hfill  
    \subfigure[Visualization Results of Sample tumor51 under Three-Class Clustering with $\boldsymbol{\beta}=(0.55,-0.59,-0.40)$. Red: cluster 1, blue: cluster 2, green: cluster 3.]{
        \includegraphics[width=0.45\textwidth]{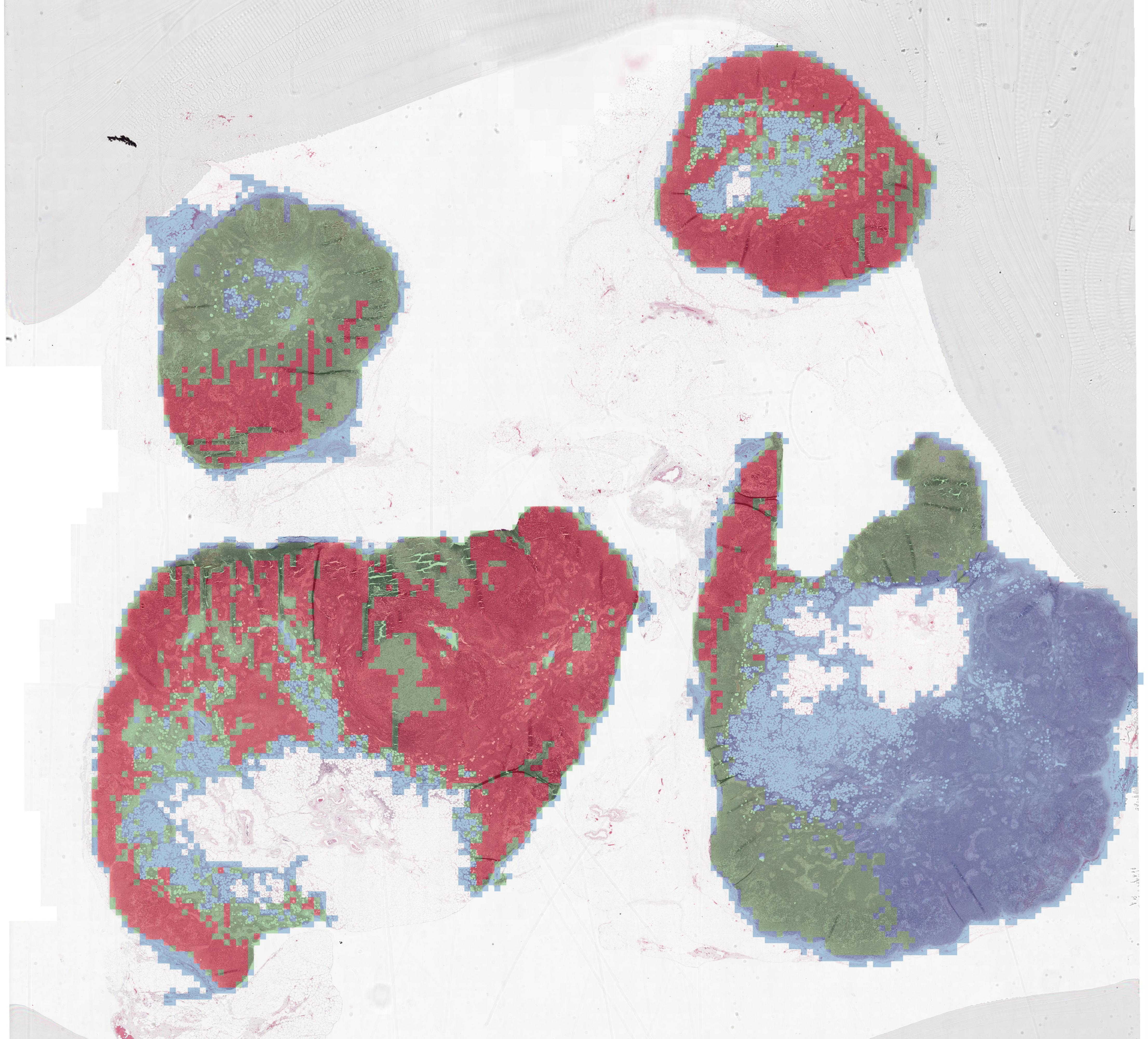}
    }
    \hfill  
    \subfigure[Visualization Results of Sample tumor51 under Five-Class Clustering with $\boldsymbol{\beta}=(0.71,0,-0.97,0.38,0.09)$, Red: cluster 1, blue: cluster 2, green: cluster 3, purple: cluster 4, yellow: cluster 5.]{
        \includegraphics[width=0.45\textwidth]{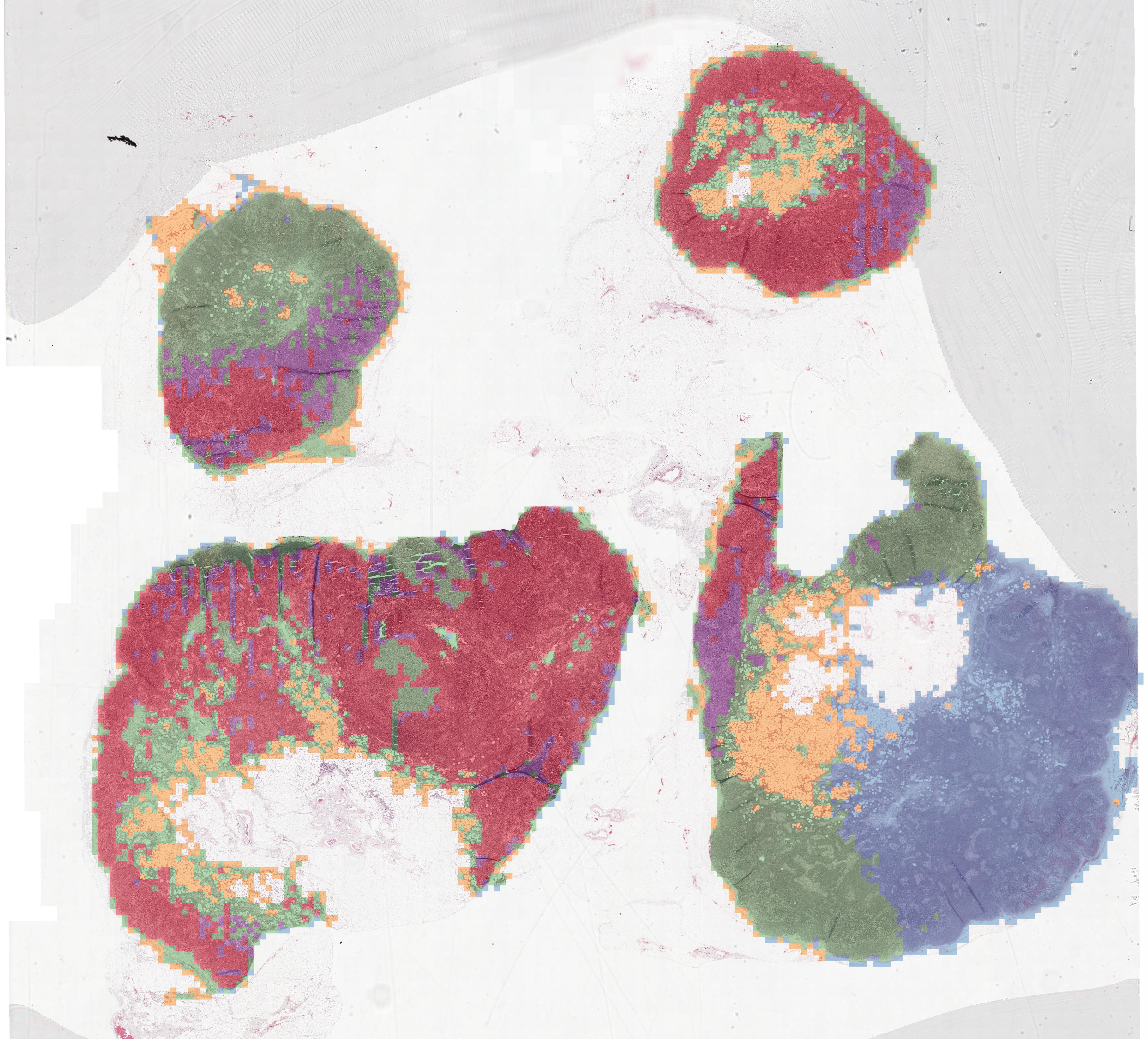}
    }
    \caption{The clustering visualization of one patient's whole-slide image.}
    \label{fig:clustering}
\end{figure*}

In this section, we attempt to interpret the model by visualizing some results. We endeavored to elucidate the clustering results and the excluded clusters from a medical perspective.
Figure \ref{fig:clustering} presents the visualization of the analysis process for a patient ``tumor51'' in the CAMELYON-16 dataset. Specifically, Figure \ref{fig:clustering} (a) shows the original whole-slide image of the patient, while Figures \ref{fig:clustering} (b), (c), and (d) correspond to the slice images of the clustering results when $K = 2$, $3$, and $5$, respectively. 

\begin{figure}
    \centering
    \includegraphics[width=1.0\linewidth]{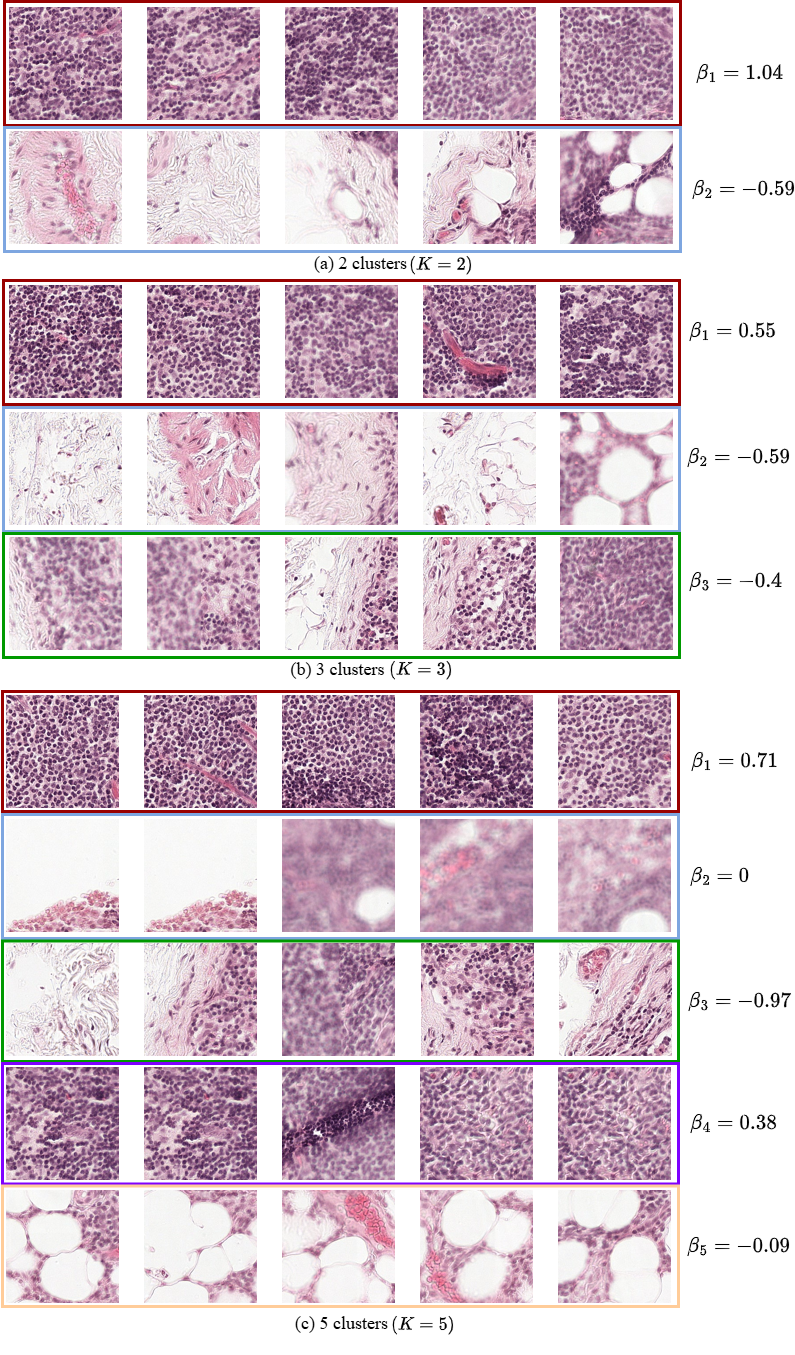}
    \caption{Visualization of patches under different clustering settings in Figure \ref{fig:clustering}.}
    \label{fig:clusteredPatches}
\end{figure}

Figure \ref{fig:clusteredPatches} displays the clustering results of some instances when $K = 2$, $3$, and $5$. In detail, Figure \ref{fig:clusteredPatches} (a) illustrates the visualization of 5 instances in two clusters when $K = 2$. Similarly, Figures \ref{fig:clusteredPatches} (b) and (c) represent the cases when $K = 3$ and $K = 5$, respectively. Noted that in Figure \ref{fig:clusteredPatches}, we use boxes of different colors to correspond to different clusters in Figure \ref{fig:clustering}. In this section, we will explain the results of the model by combining Figure \ref{fig:clustering} and Figure \ref{fig:clusteredPatches}. 

To interpret the results of the clustering and the $\ell_1$ regularization of the specific groups from a medical perspective, we analyze the results presented in Figure \ref{fig:clusteredPatches} (a)-(c), which illustrates representative instances from each cluster for settings with $K=2$, $K=3$, and $K=5$ respectively. For example, in Figure \ref{fig:clusteredPatches} (a), the model partitions all instances into two groups: five instances from the first group are highlighted within a red box, while instances from the second cluster are enclosed in a blue box. Combined with the clustering results from Figure \ref{fig:clustering}, we observe that the red cluster corresponds to a positive weight ($\beta^{K=2}_\text{red}=1.04$), whereas the blue cluster exhibits a negative weight ($\beta^{K=2}_\text{blue}=-1.59$). This indicates that the model assigns great diagnostic significance to both clusters. 
However, the instances in the red cluster show a positive correlation with the diagnostic outcomes, while the instances in the blue cluster exhibit a negative correlation with the diagnostic results.
From a medical standpoint, the red cluster predominantly contains instances characterized by dense and numerous cell nuclei (appearing as dark purple dots post-H\&E staining as shown in Figure \ref{fig:clusteredPatches}), a hallmark of tumor cell proliferation, thus warranting their retention as positive samples. Conversely, the blue cluster comprises a heterogeneous mix of patches, including tissues with larger intercellular spacing indicative of non-proliferative regions and patches containing mucus, which contribute minimally to classification outcomes.



Similarly, Figure \ref{fig:clusteredPatches} (b) visualizes partial samples from three clusters when setting $K=3$, introducing a new green cluster. Referring to Figure \ref{fig:clustering}, we observe that the majority of instances originally assigned to the red cluster under $K = 2$ remain in the red cluster under $K=3$ ($\beta^{K=3}_{\text{red}} = 0.55$), although a small subset is reassigned to the green cluster. Instances from the blue cluster under $K=2$ differentiate into two clusters under the current setting of $K=3$: the blue cluster ($\beta^{K=3}_{\text{blue}} = -0.59$) and the green cluster ($\beta^{K=3}_{\text{green}} = -0.4$). The newly formed green cluster comprises instances from both the blue and red clusters under $K=2$. Within the green cluster, many instances represent boundary regions between tissues, exhibiting both suspected proliferative cell nuclei and nuclei with normal spacing, rendering these instances ambiguous for tumor classification. Furthermore, when $K=5$, two additional clusters emerge: a purple cluster ($\beta^{K=5}_{\text{purple}} = 0.38$) and a yellow cluster ($\beta^{K=5}_{\text{yellow}} =-0.09$). Instances in the purple cluster originate from the green and red clusters of $K=3$, with the purple cluster containing instances exhibiting cell nuclei density intermediate between tumor proliferative and normal tissues. From the setting $K=3$ to $K=5$, the red cluster remains largely stable, with its weight ($\beta^{K=5}_{\text{red}} = 0.71$) underscoring its critical role in determining diagnostic outcomes. The blue cluster under $K=3$ splits into a new blue cluster ($\beta^{K=5}_{\text{blue}}=0$) and a yellow cluster ($\beta^{K=5}_{\text{yellow}} = 0.09$). At $K=5$, the yellow cluster predominantly comprises mucus instances, while the new blue cluster includes noisy patches from boundary regions and ambiguous fields of view, contributing negligibly to tumor diagnosis and thus warranting exclusion (refer to $\ell_1$ weights as shown in Figure \ref{fig:clusteredPatches}).

For CAMELYON-16, histopathology slides exhibit spatially coherent subregions (e.g., tumor vs. stroma). Moderate $K=3$ likely aligns with dominant morphological patterns, while $K=5, 10$ fragments these regions into less meaningful units. And for TCGA-NSCLC, genomic data contains latent molecular subtypes. Smaller $K$ (2 or 3) effectively groups co-expressed genes, whereas larger $K$ introduces spurious correlations from high-dimensional noise. 

\subsection{Robust of the proposed method: $\gamma$ }
\begin{figure*}[ht]
    \centering
    \subfigure[Camelyon16]{       
        \includegraphics[width=0.45\textwidth]{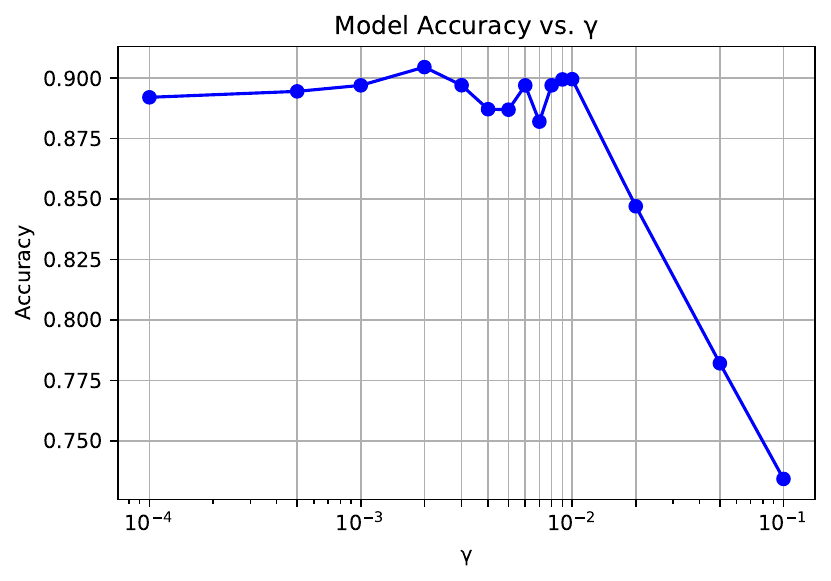}
    }
    \subfigure[TCGA-NSCLC]{        
        \includegraphics[width=0.45\textwidth]{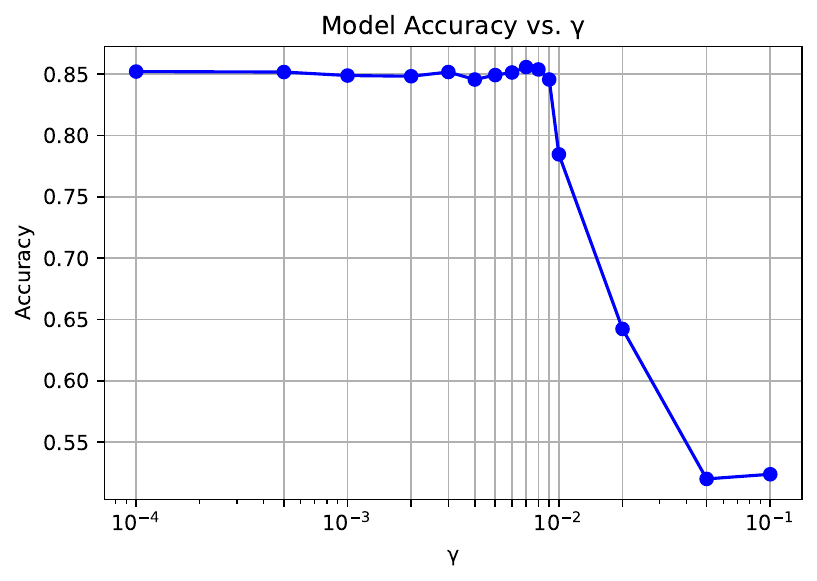}
    }
    \caption{Parameter $\gamma$ selection via network search.}
    \label{fig:gammaSearching}
\end{figure*}

In this section, we evaluate how the parameter $\gamma$ (Equation (\ref{eq:loss})) affects our model.
The analysis of model performance with respect to the parameter $\gamma$ across two distinct datasets, CAMELYON16 and TCGA-NSCLC as shown in Figure \ref{fig:gammaSearching} (a) and (b) respectively, reveals significant insights into how this parameter influences accuracy. 
The candidate set of parameter $\gamma$ is \{0.0001, 0.0005, 0.001,0.002, 0.003, 0.004, 0.005, 0.006, 0.007, 0.008,0.009, 0.01, 0.02, 0.05, 0.1 \}.

Figure \ref{fig:gammaSearching} (a) visualize search $\gamma$ with CAMELYON16 dataset. The results indicate a positive correlation between $\gamma$ values and model accuracy. As $\gamma$ increases from 0.750 to 0.900, there is a noticeable improvement in accuracy. The accuracy peaks at higher $\gamma$ levels, suggesting that tuning this parameter is crucial for optimizing model performance in this dataset.

Figure \ref{fig:gammaSearching} (b) are the process of parameter searching on TCGA-NSCLC dataset. Similar to CAMELYON16, the TCGA-NSCLC dataset also shows an increasing trend in accuracy as $\gamma$ rises from 0.55 to 0.85. However, the rate of improvement appears to plateau at higher $\gamma$ values, indicating that while higher settings can enhance accuracy, the benefits may diminish after a certain point.



\section{Conclusion}
\label{sec:conclusion}
Multi-instance learning (MIL) has gained significant traction recently due to its ability to handle weakly labeled data, making it particularly relevant in domains where precise instance-level annotations are costly or impractical.
The csMIL framework introduced in this study represents a significant advancement in addressing the challenges of MIL for weakly labeled, high-dimensional datasets, particularly in computational pathology. By integrating global-local instance clustering, within-cluster attention, and cluster-level sparsity, csMIL overcomes the limitations of traditional MIL approaches, such as instance redundancy and the lack of explicit mechanisms for discarding non-informative instances. The proposed method's ability to hierarchically organize instances into semantically meaningful clusters, coupled with sparse regularization to prioritize diagnostically relevant clusters, enhances both classification performance and interpretability. Theoretical recovery bounds underscore the efficiency of csMIL, demonstrating that it can identify relevant clusters with a minimal number of bags, aligning with principles of compressed sensing. Empirical evaluations on histopathology benchmarks, including CAMELYON16 and TCGA-NSCLC, confirm csMIL's superior performance. Furthermore, the cluster-level sparsity maps generated by csMIL provide clinically valuable insights, enabling the identification of critical regions validated by pathologists, thus bridging MIL with interpretable biomarker discovery. By unifying structured attention, clustering, and sparse learning, csMIL establishes a robust, scalable, and interpretable framework that extends the capabilities of MIL beyond instance-level heuristics. This work paves the way for broader applications in high-stakes domains, such as medical imaging and beyond, where accurate and interpretable analysis of complex datasets is paramount.



\appendix[Proof of the Theorem \ref{theorem:sparseRecovery}]
To rigorously establish the sparse recovery bound for the proposed method, we analyze it through the lens of high-dimensional statistics and compressed sensing theory. Here's a step-by-step derivation:

\begin{proof}
Problem Formulation as Sparse Linear Regression
Assume the final bag representation $ \mathbf{z}^i $ linearly predicts the label $ Y^i $:
\[
Y^i = \boldsymbol{\beta}^\top \mathbf{z}^i + \epsilon_i,
\]
where $ \epsilon_i $ is sub-Gaussian noise. Let $ \boldsymbol{\beta} \in \mathbb{R}^K $ be a sparse vector with $ s \ll K $ non-zero entries. The $\ell_1$-regularized objective becomes:
\[
\min_{\boldsymbol{\beta}} \frac{1}{2M} \sum_{i=1}^M \left(Y^i -\boldsymbol{\beta}^\top \mathbf{z}^i\right)^2 + \gamma \|\boldsymbol{\beta}\|_1.
\]

2. Key Assumptions
For exact recovery of $ \boldsymbol{\beta} $, we require:
Restricted Eigenvalue (RE) Condition:  
  There exists $ \kappa > 0 $ such that for all $ \boldsymbol{v} $ with $ \|\boldsymbol{v}\|_{0} \leq s $,
  \[
  \frac{1}{M} \sum_{i=1}^M (\boldsymbol{v}^\top \mathbf{z}^i)^2 \geq \kappa \|\boldsymbol{v}\|_2^2.
  \]
Incoherence: Columns of the design matrix $ \mathbf{Z} = [\mathbf{z}^1, ..., \mathbf{z}^M]^\top $ have bounded correlation:
  \[
  \max_{k \neq l} \left| \frac{\langle \mathbf{Z}_k, \mathbf{Z}_l \rangle}{M} \right| \leq \mu.
  \]

 3. Sparse Recovery Guarantee
Under RE and incoherence, with probability $ 1 - \delta $, the solution $ \hat{\boldsymbol{\beta}} $ satisfies:
\[
\|\hat{\boldsymbol{\beta}} \boldsymbol{\beta}^*\|_2 \leq C \sigma \sqrt{\frac{s \log K}{M}},
\]
if the number of bags satisfies:
\[
M \geq C \cdot \frac{s \log(K/\delta)}{\kappa^2 (1 \mu)^2}.
\]

 4. Interpretation of Constants
$ C $ is a universal constant (typically $ C \in [4, 16] $) derived from concentration inequalities (e.g., Hoeffding’s inequality). It depends on:
  Noise variance ($ \sigma^2 $): Higher noise increases $ C $.
  RE constant $ \kappa $ and incoherence $ \mu $: Poorer conditioning (small $ \kappa $, large $ \mu $) increases $ C $.
Practical Implications:
  Easier recovery (large $ \kappa $, small $ \mu $): Fewer bags $ M $ needed.
  Harder recovery (small $ \kappa $, large $ \mu $): Requires more bags $ M $.

 5. Connection to MIL Context
Design Matrix $ \mathbf{Z} $ corresponds to cluster-aggregated features $ \mathbf{z}^i $.
RE Condition ensures clusters are discriminative and non-redundant.
Incoherence implies clusters capture distinct aspects of the data.

 6. Limitations and Refinements
Non-linear Dependencies: The linear model assumption may not hold strictly; the bound serves as a theoretical idealization.
Data-Dependent $ C $:In practice, $ C $ is estimated via cross-validation.
Cluster Quality: Poor clustering (violating RE/incoherence) invalidates the bound.
\end{proof}

The sparse recovery bound $ M \geq C s \log(K/\delta) $ formalizes the intuition that identifying $ s $ relevant clusters out of $ K $ requires bags $ M $ scaling logarithmically with $ K $. The constant $ C $ encapsulates problem hardness, depending on data separability and noise. This aligns with compressed sensing principles, justifying the $\ell_1$ regularization’s efficacy in pruning irrelevant clusters.

Generalization Bound: With $R$-Lipschitz loss and $\|\boldsymbol{\beta}\|_1 \leq B$, Rademacher complexity scales as $O\left(BR\sqrt{\frac{K \log M}{M}}\right)$, justifying the sparsity-induced capacity control.

\section{}
Appendix two text goes here.

\ifCLASSOPTIONcompsoc
  \section{Acknowledgments}
\else
  \section{Acknowledgment}
\fi

The authors would like to thank...

\ifCLASSOPTIONcaptionsoff
  \newpage
\fi


\bibliographystyle{IEEEtran}
\bibliography{reference}

\begin{thebibliography}{10}
\providecommand{\url}[1]{#1}
\csname url@samestyle\endcsname
\providecommand{\newblock}{\relax}
\providecommand{\bibinfo}[2]{#2}
\providecommand{\BIBentrySTDinterwordspacing}{\spaceskip=0pt\relax}
\providecommand{\BIBentryALTinterwordstretchfactor}{4}
\providecommand{\BIBentryALTinterwordspacing}{\spaceskip=\fontdimen2\font plus
\BIBentryALTinterwordstretchfactor\fontdimen3\font minus
  \fontdimen4\font\relax}
\providecommand{\BIBforeignlanguage}[2]{{%
\expandafter\ifx\csname l@#1\endcsname\relax
\typeout{** WARNING: IEEEtran.bst: No hyphenation pattern has been}%
\typeout{** loaded for the language `#1'. Using the pattern for}%
\typeout{** the default language instead.}%
\else
\language=\csname l@#1\endcsname
\fi
#2}}
\providecommand{\BIBdecl}{\relax}
\BIBdecl

\bibitem{lu2021data}
M.~Y. Lu, D.~F. Williamson, T.~Y. Chen, R.~J. Chen, M.~Barbieri, and
  F.~Mahmood, ``Data-efficient and weakly supervised computational pathology on
  whole-slide images,'' \emph{Nature Biomedical Engineering}, vol.~5, no.~6,
  pp. 555--570, 2021.

\bibitem{GADERMAYR2024102337}
\BIBentryALTinterwordspacing
M.~Gadermayr and M.~Tschuchnig, ``Multiple instance learning for digital
  pathology: A review of the state-of-the-art, limitations \& future
  potential,'' \emph{Computerized Medical Imaging and Graphics}, vol. 112, p.
  102337, 2024. [Online]. Available:
  \url{https://www.sciencedirect.com/science/article/pii/S0895611124000144}
\BIBentrySTDinterwordspacing

\bibitem{liu2024pseudo}
P.~Liu, L.~Ji, X.~Zhang, and F.~Ye, ``Pseudo-bag mixup augmentation for
  multiple instance learning-based whole slide image classification,''
  \emph{IEEE Transactions on Medical Imaging}, vol.~43, no.~5, pp. 1841--1852,
  2024.

\bibitem{liu2024advmil}
P.~Liu, L.~Ji, F.~Ye, and B.~Fu, ``Advmil: Adversarial multiple instance
  learning for the survival analysis on whole-slide images,'' \emph{Medical
  Image Analysis}, vol.~91, p. 103020, 2024.

\bibitem{maron1997framework}
O.~Maron and T.~Lozano-P{\'e}rez, ``A framework for multiple-instance
  learning,'' \emph{Advances in neural information processing systems},
  vol.~10, 1997.

\bibitem{andrews2002support}
S.~Andrews, I.~Tsochantaridis, and T.~Hofmann, ``Support vector machines for
  multiple-instance learning,'' \emph{Advances in neural information processing
  systems}, vol.~15, 2002.

\bibitem{sun2016multiple}
M.~Sun, T.~X. Han, M.-C. Liu, and A.~Khodayari-Rostamabad, ``Multiple instance
  learning convolutional neural networks for object recognition,'' in
  \emph{2016 23rd International Conference on Pattern Recognition
  (ICPR)}.\hskip 1em plus 0.5em minus 0.4em\relax IEEE, 2016, pp. 3270--3275.

\bibitem{ilse2018icml}
M.~Ilse, J.~Tomczak, and M.~Welling, ``Attention-based deep multiple instance
  learning,'' in \emph{Proceedings of the 35th International Conference on
  Machine Learning}, ser. Proceedings of Machine Learning Research, J.~Dy and
  A.~Krause, Eds., vol.~80.\hskip 1em plus 0.5em minus 0.4em\relax PMLR, 10--15
  Jul 2018, pp. 2127--2136.

\bibitem{shi2020loss}
X.~Shi, F.~Xing, Y.~Xie, Z.~Zhang, L.~Cui, and L.~Yang, ``Loss-based attention
  for deep multiple instance learning,'' in \emph{Proceedings of the AAAI
  conference on artificial intelligence}, vol.~34, no.~04, 2020, pp.
  5742--5749.

\bibitem{tang2024multi}
W.~Tang, Y.-F. Yang, Z.~Wang, W.~Zhang, and M.-L. Zhang, ``Multi-instance
  partial-label learning with margin adjustment,'' \emph{Advances in Neural
  Information Processing Systems}, vol.~37, pp. 26\,331--26\,354, 2024.

\bibitem{chen2006pami}
Y.~Chen, J.~Bi, and J.~Wang, ``Miles: Multiple-instance learning via embedded
  instance selection,'' \emph{IEEE Transactions on Pattern Analysis and Machine
  Intelligence}, vol.~28, no.~12, pp. 1931--1947, 2006.

\bibitem{sofiiuk2019adaptis}
K.~Sofiiuk, O.~Barinova, and A.~Konushin, ``Adaptis: Adaptive instance
  selection network,'' in \emph{Proceedings of the IEEE/CVF international
  conference on computer vision}, 2019, pp. 7355--7363.

\bibitem{li2021deep}
X.-C. Li, D.-C. Zhan, J.-Q. Yang, and Y.~Shi, ``Deep multiple instance
  selection,'' \emph{Science China Information Sciences}, vol.~64, pp. 1--15,
  2021.

\bibitem{zhang2022sparse}
Y.~Zhang, D.~Su, X.~Zhao, and Y.~Tian, ``Sparse multiple instance learning for
  elderly people balance ability,'' \emph{Procedia Computer Science}, vol. 199,
  pp. 621--628, 2022.

\bibitem{dundr2021primary}
P.~Dundr, N.~Singh, B.~No{\v{z}}i{\v{c}}kov{\'a}, K.~N{\v{e}}mejcov{\'a},
  M.~B{\'a}rtů, and I.~Stru{\v{z}}insk{\'a}, ``Primary mucinous ovarian tumors
  vs. ovarian metastases from gastrointestinal tract, pancreas and biliary
  tree: a review of current problematics,'' \emph{Diagnostic pathology},
  vol.~16, pp. 1--17, 2021.

\bibitem{leen2012pathology}
S.~L.~S. Leen and N.~Singh, ``Pathology of primary and metastatic mucinous
  ovarian neoplasms,'' \emph{Journal of Clinical Pathology}, vol.~65, no.~7,
  pp. 591--595, 2012.

\bibitem{zhang2022dtfd}
H.~Zhang, Y.~Meng, Y.~Zhao, Y.~Qiao, X.~Yang, S.~E. Coupland, and Y.~Zheng,
  ``Dtfd-mil: Double-tier feature distillation multiple instance learning for
  histopathology whole slide image classification,'' in \emph{Proceedings of
  the IEEE/CVF conference on computer vision and pattern recognition}, 2022,
  pp. 18\,802--18\,812.

\bibitem{campanella2019clinical}
G.~Campanella, M.~G. Hanna, L.~Geneslaw, A.~Miraflor, V.~Werneck Krauss~Silva,
  K.~J. Busam, E.~Brogi, V.~E. Reuter, D.~S. Klimstra, and T.~J. Fuchs,
  ``Clinical-grade computational pathology using weakly supervised deep
  learning on whole slide images,'' \emph{Nature Medicine}, vol.~25, no.~8, pp.
  1301--1309, 2019.

\bibitem{dietterich1997solving}
T.~G. Dietterich, R.~H. Lathrop, and T.~Lozano-P{\'e}rez, ``Solving the
  multiple instance problem with axis-parallel rectangles,'' \emph{Artificial
  intelligence}, vol.~89, no. 1-2, pp. 31--71, 1997.

\bibitem{ramon2000multi}
J.~Ramon and L.~De~Raedt, ``Multi instance neural networks,'' in
  \emph{Proceedings of the ICML-2000 workshop on attribute-value and relational
  learning}, 2000, pp. 53--60.

\bibitem{liu2012key}
G.~Liu, J.~Wu, and Z.-H. Zhou, ``Key instance detection in multi-instance
  learning,'' in \emph{Asian conference on machine learning}.\hskip 1em plus
  0.5em minus 0.4em\relax PMLR, 2012, pp. 253--268.

\bibitem{carbonneau2018multiple}
M.-A. Carbonneau, V.~Cheplygina, E.~Granger, and G.~Gagnon, ``Multiple instance
  learning: A survey of problem characteristics and applications,''
  \emph{Pattern recognition}, vol.~77, pp. 329--353, 2018.

\bibitem{kotzias2015group}
D.~Kotzias, M.~Denil, N.~De~Freitas, and P.~Smyth, ``From group to individual
  labels using deep features,'' in \emph{Proceedings of the 21th ACM SIGKDD
  international conference on knowledge discovery and data mining}, 2015, pp.
  597--606.

\bibitem{Andrews2002MISVM}
S.~Andrews, I.~Tsochantaridis, and T.~Hofmann, ``Support vector machines for
  multiple-instance learning,'' in \emph{Proceedings of the 16th International
  Conference on Neural Information Processing Systems}, ser. NIPS'02.\hskip 1em
  plus 0.5em minus 0.4em\relax Cambridge, MA, USA: MIT Press, 2002, p.
  577–584.

\bibitem{zhang2001dd}
Q.~Zhang and S.~Goldman, ``Em-dd: An improved multiple-instance learning
  technique,'' \emph{Advances in neural information processing systems},
  vol.~14, 2001.

\bibitem{zaheer2017deep}
M.~Zaheer, S.~Kottur, S.~Ravanbakhsh, B.~Poczos, R.~R. Salakhutdinov, and A.~J.
  Smola, ``Deep sets,'' \emph{Advances in neural information processing
  systems}, vol.~30, 2017.

\bibitem{wu2015deep}
J.~Wu, Y.~Yu, C.~Huang, and K.~Yu, ``Deep multiple instance learning for image
  classification and auto-annotation,'' in \emph{Proceedings of the IEEE
  conference on computer vision and pattern recognition}, 2015, pp. 3460--3469.

\bibitem{shao2021transmil}
Z.~Shao, H.~Bian, Y.~Chen, Y.~Wang, J.~Zhang, X.~Ji \emph{et~al.},
  ``Trans{MIL}: Transformer based correlated multiple instance learning for
  whole slide image classification,'' \emph{Advances in neural information
  processing systems}, vol.~34, pp. 2136--2147, 2021.

\bibitem{li2017pami}
B.~Li, C.~Yuan, W.~Xiong, W.~Hu, H.~Peng, X.~Ding, and S.~Maybank, ``Multi-view
  multi-instance learning based on joint sparse representation and multi-view
  dictionary learning,'' \emph{IEEE Transactions on Pattern Analysis and
  Machine Intelligence}, vol.~39, no.~12, pp. 2554--2560, 2017.

\bibitem{cinbis2016weakly}
R.~G. Cinbis, J.~Verbeek, and C.~Schmid, ``Weakly supervised object
  localization with multi-fold multiple instance learning,'' \emph{IEEE
  transactions on pattern analysis and machine intelligence}, vol.~39, no.~1,
  pp. 189--203, 2016.

\bibitem{gu2008multi}
Z.~Gu, T.~Mei, X.-S. Hua, J.~Tang, and X.~Wu, ``Multi-layer multi-instance
  learning for video concept detection,'' \emph{IEEE Transactions on
  Multimedia}, vol.~10, no.~8, pp. 1605--1616, 2008.

\bibitem{shao2023video}
W.~Shao, R.~Xiao, P.~Rajapaksha, M.~Wang, N.~Crespi, Z.~Luo, and R.~Minerva,
  ``Video anomaly detection with ntcn-ml: A novel tcn for multi-instance
  learning,'' \emph{Pattern Recognition}, vol. 143, p. 109765, 2023.

\bibitem{he2009text}
W.~He and Y.~Wang, ``Text representation and classification based on
  multi-instance learning,'' in \emph{2009 International Conference on
  Management Science and Engineering}.\hskip 1em plus 0.5em minus 0.4em\relax
  IEEE, 2009, pp. 34--39.

\bibitem{tong2014multiple}
T.~Tong, R.~Wolz, Q.~Gao, R.~Guerrero, J.~V. Hajnal, D.~Rueckert, A.~D.~N.
  Initiative \emph{et~al.}, ``Multiple instance learning for classification of
  dementia in brain mri,'' \emph{Medical image analysis}, vol.~18, no.~5, pp.
  808--818, 2014.

\bibitem{courtiol2018classification}
P.~Courtiol, E.~W. Tramel, M.~Sanselme, and G.~Wainrib, ``Classification and
  disease localization in histopathology using only global labels: A
  weakly-supervised approach,'' \emph{arXiv preprint arXiv:1802.02212}, 2018.

\bibitem{yao2020whole}
J.~Yao, X.~Zhu, J.~Jonnagaddala, N.~Hawkins, and J.~Huang, ``Whole slide images
  based cancer survival prediction using attention guided deep multiple
  instance learning networks,'' \emph{Medical image analysis}, vol.~65, p.
  101789, 2020.

\bibitem{li2021dual}
B.~Li, Y.~Li, and K.~W. Eliceiri, ``Dual-stream multiple instance learning
  network for whole slide image classification with self-supervised contrastive
  learning,'' in \emph{Proceedings of the IEEE/CVF conference on computer
  vision and pattern recognition}, 2021, pp. 14\,318--14\,328.

\bibitem{zhu2022murcl}
Z.~Zhu, L.~Yu, W.~Wu, R.~Yu, D.~Zhang, and L.~Wang, ``Murcl: Multi-instance
  reinforcement contrastive learning for whole slide image classification,''
  \emph{IEEE Transactions on Medical Imaging}, vol.~42, no.~5, pp. 1337--1348,
  2022.

\bibitem{wang2024advances}
J.~Wang, Y.~Mao, N.~Guan, and C.~J. Xue, ``Advances in multiple instance
  learning for whole slide image analysis: Techniques, challenges, and future
  directions,'' \emph{arXiv preprint arXiv:2408.09476}, 2024.

\bibitem{cai2024rethinking}
L.~Cai, S.~Huang, Y.~Zhang, J.~Lu, and Y.~Zhang, ``Rethinking attention-based
  multiple instance learning for whole-slide pathological image classification:
  An instance attribute viewpoint,'' \emph{arXiv preprint arXiv:2404.00351},
  2024.

\bibitem{zhang2024aem}
Y.~Zhang, Z.~Shui, Y.~Sun, H.~Li, J.~Li, C.~Zhu, and L.~Yang, ``Aem: Attention
  entropy maximization for multiple instance learning based whole slide image
  classification,'' \emph{arXiv preprint arXiv:2406.15303}, 2024.

\bibitem{qu2024rethinking}
L.~Qu, Y.~Ma, X.~Luo, Q.~Guo, M.~Wang, and Z.~Song, ``Rethinking multiple
  instance learning for whole slide image classification: A good instance
  classifier is all you need,'' \emph{IEEE Transactions on Circuits and Systems
  for Video Technology}, vol.~34, no.~10, pp. 9732--9744, 2024.

\bibitem{liu2024attention}
X.~Liu, W.~Zhang, and M.-L. Zhang, ``Attention is not what you need: Revisiting
  multi-instance learning for whole slide image classification,'' \emph{arXiv
  preprint arXiv:2408.09449}, 2024.

\bibitem{10810475}
C.~Jin, L.~Luo, H.~Lin, J.~Hou, and H.~Chen, ``Hmil: Hierarchical
  multi-instance learning for fine-grained whole slide image classification,''
  \emph{IEEE Transactions on Medical Imaging}, vol.~44, no.~4, pp. 1796--1808,
  2025.

\bibitem{mao2025camil}
J.~Mao, J.~Xu, X.~Tang, Y.~Liu, H.~Zhao, G.~Tian, and J.~Yang, ``Camil: channel
  attention-based multiple instance learning for whole slide image
  classification,'' \emph{Bioinformatics}, vol.~41, no.~2, p. btaf024, 2025.

\bibitem{wang2023targeting}
Z.~Wang, Y.~Bi, T.~Pan, X.~Wang, C.~Bain, R.~Bassed, S.~Imoto, J.~Yao, R.~J.
  Daly, and J.~Song, ``Targeting tumor heterogeneity: multiplex-detection-based
  multiple instance learning for whole slide image classification,''
  \emph{Bioinformatics}, vol.~39, no.~3, p. btad114, 2023.

\bibitem{chen2024camil}
K.~Chen, S.~Sun, and J.~Zhao, ``Camil: Causal multiple instance learning for
  whole slide image classification,'' in \emph{Proceedings of the AAAI
  Conference on Artificial Intelligence}, vol.~38, no.~2, 2024, pp. 1120--1128.

\bibitem{keshvarikhojasteh2024multiple}
H.~Keshvarikhojasteh, J.~P. Pluim, and M.~Veta, ``Multiple instance learning
  with random sampling for whole slide image classification,'' in \emph{Medical
  Imaging 2024: Digital and Computational Pathology}, vol. 12933.\hskip 1em
  plus 0.5em minus 0.4em\relax SPIE, 2024, pp. 372--376.

\bibitem{5357420}
B.~Cheng, J.~Yang, S.~Yan, Y.~Fu, and T.~S. Huang, ``Learning with $\ell
  ^{1}$-graph for image analysis,'' \emph{IEEE Transactions on Image
  Processing}, vol.~19, no.~4, pp. 858--866, 2010.

\bibitem{zhou2009icml}
Z.-H. Zhou, Y.-Y. Sun, and Y.-F. Li, ``Multi-instance learning by treating
  instances as non-i.i.d. samples,'' in \emph{Proceedings of the 26th Annual
  International Conference on Machine Learning}, ser. ICML '09.\hskip 1em plus
  0.5em minus 0.4em\relax New York, NY, USA: Association for Computing
  Machinery, 2009, p. 1249–1256.

\bibitem{li2021adaptive}
G.~Li, V.~Jampani, L.~Sevilla-Lara, D.~Sun, J.~Kim, and J.~Kim, ``Adaptive
  prototype learning and allocation for few-shot segmentation,'' in
  \emph{Proceedings of the IEEE/CVF conference on computer vision and pattern
  recognition}, 2021, pp. 8334--8343.

\bibitem{candes2008enhancing}
E.~J. Candes, M.~B. Wakin, and S.~P. Boyd, ``Enhancing sparsity by reweighted
  l\_1 minimization,'' \emph{Journal of Fourier analysis and applications},
  vol.~14, pp. 877--905, 2008.

\bibitem{bejnordi2017diagnostic}
B.~E. Bejnordi, M.~Veta, P.~J. Van~Diest, B.~Van~Ginneken, N.~Karssemeijer,
  G.~Litjens, J.~A. Van Der~Laak, M.~Hermsen, Q.~F. Manson, M.~Balkenhol
  \emph{et~al.}, ``Diagnostic assessment of deep learning algorithms for
  detection of lymph node metastases in women with breast cancer,''
  \emph{Jama}, vol. 318, no.~22, pp. 2199--2210, 2017.

\bibitem{cancer2012comprehensive}
C.~G.~A. Network, ``Comprehensive molecular portraits of human breast
  tumours,'' \emph{Nature}, vol. 490, no. 7418, pp. 61--70, 2012.

\bibitem{lee2019set}
J.~Lee, Y.~Lee, J.~Kim, A.~Kosiorek, S.~Choi, and Y.~W. Teh, ``Set transformer:
  A framework for attention-based permutation-invariant neural networks,'' in
  \emph{International conference on machine learning}.\hskip 1em plus 0.5em
  minus 0.4em\relax PMLR, 2019, pp. 3744--3753.

\end{thebibliography}

\end{document}